\documentclass[10pt,twocolumn,letterpaper]{article}

\usepackage[pagenumbers]{cvpr} %

\usepackage{amsmath}
\usepackage{amssymb}
\usepackage{booktabs}

\usepackage{xcolor,color,framed}
\usepackage{url, overpic, cases, contour}
\usepackage{multirow, makecell}
\newcommand{\R}[1]{\textcolor[rgb]{1.00,0.00,0.00}{#1}}
\newcommand{\B}[1]{\textcolor[rgb]{0.00,0.00,1.00}{#1}}
\definecolor{shadecolor}{rgb}{0.92,0.92,0.92}

\usepackage{adjustbox}
\makeatletter %
\@namedef{ver@everyshi.sty}{}
\makeatother
\usepackage{tikz}

\newcommand{\name}{0}
\newcommand{\h}{0}
\newcommand{\w}{0.15}
\newcommand{\wa}{0.15}
\newlength \g

\usepackage[pagebackref,breaklinks,colorlinks]{hyperref}
\usepackage[symbol]{footmisc}

\usepackage[capitalize]{cleveref}
\crefname{section}{Sec.}{Secs.}
\Crefname{section}{Section}{Sections}
\Crefname{table}{Table}{Tables}
\crefname{table}{Tab.}{Tabs.}

\def\cvprPaperID{} %
\def\confName{CVPR}
\def\confYear{2022}

\begin{document}

\title{VRT: A Video Restoration Transformer}
\author{Jingyun Liang$^{1}$ \qquad Jiezhang Cao$^{1}$ \qquad Yuchen Fan$^{2}$  \qquad Kai Zhang$^{1}$\thanks{Corresponding author.}\\ Rakesh Ranjan$^{2}$\qquad  \qquad Yawei Li$^{1}$ \qquad Radu Timofte$^{1}$ \qquad Luc Van Gool$^{1,3}$ \\
$^{1}$ Computer Vision Lab, ETH Zurich, Switzerland \quad
$^{2}$ Meta Inc. \quad $^{3}$ KU Leuven, Belgium\\
{\tt\small \{jinliang, jiezcao, kaizhang, vangool, timofter\}@vision.ee.ethz.ch \quad\{ycfan, rakeshr\}@fb.com}\quad 
{\tt\small }\\
\url{https://github.com/JingyunLiang/VRT}
}
\maketitle

\begin{abstract}
Video restoration (\eg, video super-resolution) aims to restore high-quality frames from low-quality frames. Different from single image restoration, video restoration generally requires to utilize temporal information from multiple adjacent but usually misaligned video frames. Existing deep methods generally tackle with this by exploiting a sliding window strategy or a recurrent architecture, which either is restricted by frame-by-frame restoration or lacks long-range modelling ability. In this paper, we propose a Video Restoration Transformer (VRT) with parallel frame prediction and long-range temporal dependency modelling abilities. More specifically, VRT is composed of multiple scales, each of which consists of two kinds of modules: temporal mutual self attention (TMSA) and parallel warping. TMSA divides the video into small clips, on which mutual attention is applied for joint motion estimation, feature alignment and feature fusion, while self attention is used for feature extraction. To enable cross-clip interactions, the video sequence is shifted for every other layer. Besides, parallel warping is used to further fuse information from neighboring frames by parallel feature warping. Experimental results on five tasks, including video super-resolution, video deblurring, video denoising, video frame interpolation and space-time video super-resolution, demonstrate that VRT outperforms the state-of-the-art methods by large margins (\textbf{up to 2.16dB}) on fourteen benchmark datasets.
\end{abstract}

\section{Introduction}
Video restoration, which reconstructs high-quality (HQ) frames from multiple low-quality (LQ) frames, has attracted much attention recently. Compared with image restoration, the key challenge of video restoration lies in how to make full use of neighboring highly-related but misaligned supporting frames for reconstructing reference frames. 

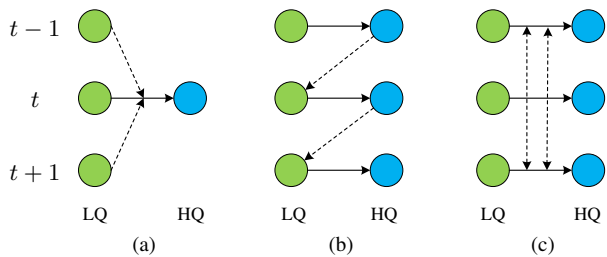
\begin{figure}
\captionsetup{font=small}%
\scriptsize
\hspace{0.35cm}
 \begin{subfigure}{.15\textwidth}
   \centering
   \begin{overpic}[width=0.66\textwidth]{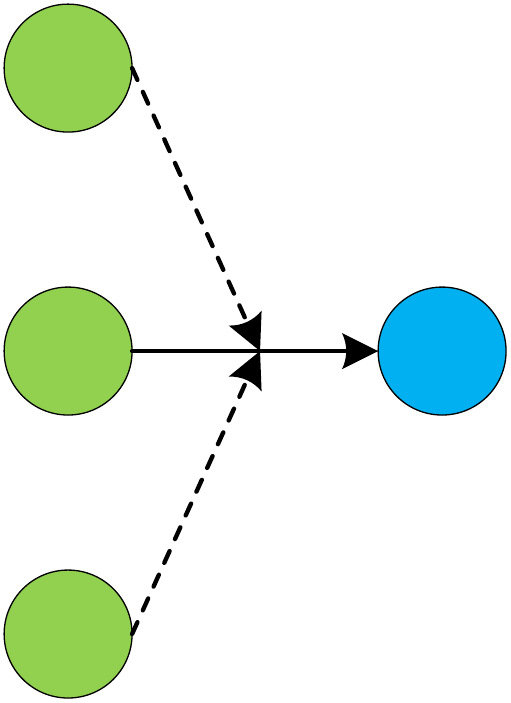}
     \put(-38,85){\color{black}{\small ${t-1}$}}
     \put(-26,45){\color{black}{\small ${t}$}}
     \put(-38,4){\color{black}{\small ${t+1}$}}
     \put(3,-18){\color{black}{\scriptsize LQ}}
     \put(56,-18){\color{black}{\scriptsize HQ}}
     \put(114,-18){\color{black}{\scriptsize LQ}}
     \put(164,-18){\color{black}{\scriptsize HQ}}
     \put(225.5,-18){\color{black}{\scriptsize LQ}}
     \put(277.5,-18){\color{black}{\scriptsize HQ}}
     \end{overpic}
   \vspace{0.5cm}
   \caption{}
   \label{fig:intro_sliding}
 \end{subfigure}%
 \begin{subfigure}{.15\textwidth}
   \centering
   \includegraphics[width=.66\linewidth]{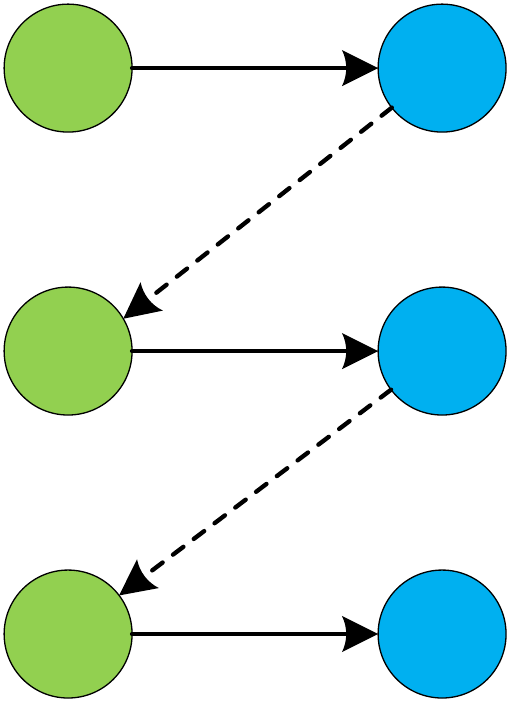}
   \vspace{0.5cm}
   \caption{}
   \label{fig:intro_rnn}
 \end{subfigure}
 \begin{subfigure}{.15\textwidth}
   \centering
   \includegraphics[width=.66\linewidth]{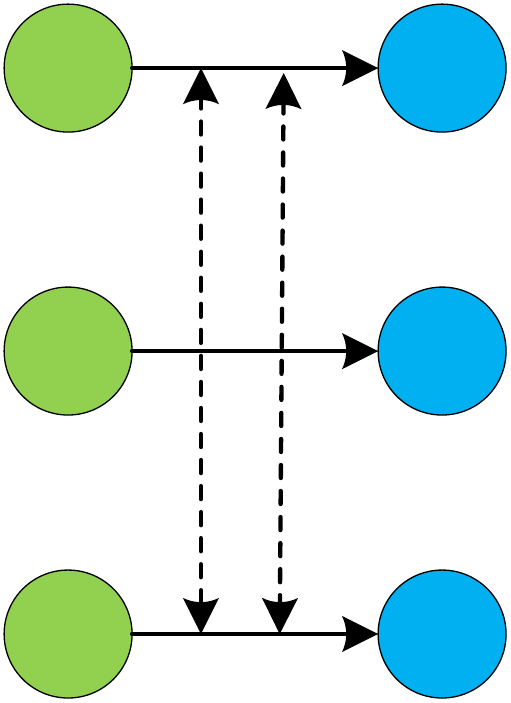}
   \vspace{0.5cm}
   \caption{}
   \label{fig:intro_vrt}
 \end{subfigure}
 \vspace{-0.1cm}
 \caption{Illustrative comparison of sliding window-based models (\ref{fig:intro_sliding}, \eg,~\cite{li2020mucan, tian2020tdan, wang2019edvr}), recurrent models (\ref{fig:intro_rnn}, \eg,~\cite{fuoli2019rlsp, huang2015bidirectional, isobe2020rsdn, chan2021basicvsr, chan2021basicvsr++}) and the proposed parallel VRT model (\ref{fig:intro_vrt}). Green and blue circles denote low-quality (LQ) input frames and high-quality (HQ) output frames, respectively. $t-1$, $t$ and $t+1$ are frame serial numbers. Dashed lines represent information fusion among different frames.}
 \label{fig:intro}
 \vspace{-0.2cm}
\end{figure}

Existing video restoration methods can be mainly divided into two categories: sliding window-based methods~\cite{caballero2017VESPCN, huang2017video, wang2019edvr, tian2020tdan, li2020mucan, su2017dvddeblur, zhou2019spatio, isobe2020tga, li2021arvo} and recurrent methods~\cite{huang2015bidirectional, sajjadi2018FRVSR, fuoli2019rlsp, haris2019recurrent, isobe2020rsdn, isobe2020rrn, chan2021basicvsr, chan2021basicvsr++, lin2021fdan, nah2019recurrent, zhong2020efficient, son2021recurrent}. As shown in Fig.~\ref{fig:intro_sliding}, sliding window-based methods generally input multiple frames to generate a single HQ frame and processes long video sequences in a sliding window fashion. Each input frame is processed for multiple times in inference, leading to inefficient feature utilization and increased computation cost. 

Some other methods are based on a recurrent architecture. As shown in Fig.~\ref{fig:intro_rnn}, recurrent models mainly use previously reconstructed HQ frames for subsequent frame reconstruction. Due to the recurrent nature, they have three disadvantages. First, recurrent methods are limited in parallelization for efficient distributed training and inference. Second, although information is accumulated frame by frame, recurrent models are not good at long-range temporal dependency modelling. One frame may strongly affect the next adjacent frame, but its influence is quickly lost after few time steps~\cite{greaves2019statistical, vaswani2017transformer}. Third, they suffer from significant performance drops on few-frame videos~\cite{cao2021videosr}.

In this paper, we propose a Video Restoration Transformer (VRT) that allows for parallel computation and long-range dependency modelling in video restoration. Based on a multi-scale framework, VRT divides the video sequence into non-overlapping clips and shifts it alternately to enable inter-clip interactions. Specifically, each scale of VRT has several temporal mutual self attention (TMSA) modules followed by a parallel warping module. In TMSA, mutual attention is focused on mutual alignment between neighboring two-frame clips, while self attention is used for feature extraction. At the end of each scale, we further use parallel warping to fuse neighboring frame information into the current frame. After multi-scale feature extraction, alignment and fusion, the HQ frames are individually reconstructed from their corresponding frame features. 

Compared with existing video restoration frameworks, VRT has several benefits. First, as shown in Fig.~\ref{fig:intro_vrt}, VRT is trained and tested on long video sequences in parallel. In contrast, both sliding window-based and recurrent methods are often tested frame by frame. Second, VRT has the ability to model long-range temporal dependencies, utilizing information from multiple neighbouring frames during the reconstruction of each frame. By contrast, sliding window-based methods cannot be easily scaled up to long sequence modelling, while recurrent methods may forget distant information after several timestamps. Third, VRT proposes to use mutual attention for joint feature alignment and fusion. It adaptively utilizes features from supporting frames and fuses them into the reference frame, which can be regarded as implicit motion estimation and feature warping. 

Our contributions can be summarized as follows:
\begin{itemize}
  \vspace{-0.2cm}
  \item[1)] We propose a new framework named Video Restoration Transformer (VRT) that is characterized by parallel computation and long-range dependency modelling. It jointly extracts, aligns, and fuses frame features at multiple scales.
  \vspace{-0.2cm}
  \item[2)] We propose the mutual attention for mutual alignment between frames. It is a generalized ``soft'' version of image warping after implicit motion estimation.
  \vspace{-0.2cm}
  \item[3)] VRT achieves state-of-the-art performance on video restoration, including video super-resolution, deblurring, denoising, frame interpolation and space-time video super-resolution. It outperforms state-of-the-art methods by up to 2.16dB on benchmark datasets.
\end{itemize}

\section{Related Work}
\subsection{Video Restoration}
Similar to image restoration~\cite{dong2014srcnn, zhang2017DnCNN, ledig2017srresnet, fan2017balanced, liu2018NLRN, yu2018wide, zhang2018RDN, zhang2018srmd, zhang2018rcan,zhang2019RNAN, wang2019learning,mei2020CSNLN, guo2020drn,fan2020scale, liang2021fkp, xiang2021boosting, liang21manet, mei2021NLSA, liang21hcflow, wang2021unsupervised, wang2021learning,kai2021bsrgan, li2021beginner, zhang2021DPIR, sun2021mefnet, liang21swinir}, learning-based methods, especially CNN-based methods, have become the primary workhorse for video restoration~\cite{liu2018learning, wang2019edvr, zhou2019spatio, yi2019pfnl_udm, xiang2020deep, xiang2020zooming, pan2020tspdeblur, wang2020deep, pan2020cascaded, yi2021omniscient, chan2021basicvsr, maggioni2021efficient, vaksman2021patch, lee2021restore, yang2021ntire}. 

\vspace{-0.4cm}
\paragraph{Framework design.}
From the perspective of architecture design, existing methods can be roughly divided into two categories: sliding window-based and recurrent methods. Sliding window-based methods often takes a short sequence of frames as input and merely predict the center frame~\cite{caballero2017VESPCN, huang2017video, wang2019edvr, tassano2019dvdnet, tian2020tdan, li2020mucan, su2017dvddeblur, zhou2019spatio, isobe2020tga,tassano2020fastdvdnet, sheth2021unsupervised, li2021arvo}. Although some works~\cite{li2019fast} predict multiple frames, they still focus on the reconstruction of the center frame during training and testing. Recurrent framework is another popular choice~\cite{huang2015bidirectional, sajjadi2018FRVSR, fuoli2019rlsp, haris2019recurrent, isobe2020rsdn, isobe2020rrn, chan2021basicvsr, chan2021basicvsr++, lin2021fdan, nah2019recurrent, zhong2020efficient, son2021recurrent}. Huang~\etal~\cite{huang2015bidirectional} propose a bidirectional recurrent convolutional neural network for SR. Sajjadi~\etal~\cite{sajjadi2018FRVSR} warp the previous frame prediction onto the current frame and feed it to a restoration network along with the current input frame. This idea is used by Chan~\etal~\cite{chan2021basicvsr} for bidirectional recurrent network, and further extended as grid propagation in~\cite{chan2021basicvsr++}. 

\vspace{-0.4cm}
\paragraph{Temporal alignment and fusion.}
Since supporting frames are often highly-related but misaligned, temporal alignment plays an critical role in video restoration~\cite{liao2015video, xue2019TOFlow-Vimeo-90K, tian2020tdan, wang2019edvr, chan2021understanding, chan2021basicvsr, chan2021basicvsr++}. Early methods~\cite{liao2015video, kappeler2016video, caballero2017VESPCN, liu2017robust, tao2017detail} use traditional flow estimation methods to estimate optical flow and warp the supporting frames towards the reference frame. To compensate occlusion and large motion, Xue~\etal~\cite{xue2019TOFlow-Vimeo-90K} utilize task-oriented flow by fine-tuning the pre-trained optical flow estimation model SpyNet~\cite{ranjan2017spynet} on different video restoration tasks. Jo~\etal~\cite{jo2018DUF} use dynamic upsampling filters for implicit motion compensation. Kim~\etal~\cite{kim2018spatio} propose a spatio-temporal transformer network for multi-frame optical flow estimation and warping. Tian~\etal~\cite{tian2020tdan} propose TDAN that utilize deformable convolution~\cite{dai2017deformable} for feature alignment. Based on TDAN, Wang~\etal~\cite{wang2019edvr} extend it to multi-scale alignment, while Chan~\etal~\cite{chan2021basicvsr++} incorporate optical flow as a guidance for offsets learning. 

\vspace{-0.4cm}
\paragraph{Attention mechanism.}
Attention mechanism has been exploited in video restoration in combination with CNN~\cite{liu2017robust, wang2019edvr, suin2021gated, cao2021videosr}. Liu~\etal~\cite{liu2017robust} learn different weights for different temporal branches. Wang~\etal~\cite{wang2019edvr} learn pixel-level attention maps for spatial and temporal feature fusion. To better incorporate temporal information, Isobe~\etal~\cite{isobe2020tga} divide frames into several groups and design a temporal group attention module. Suin~\etal~\cite{suin2021gated} propose a reinforcement learning-based framework with factorized spatio-temporal attention. Cao~\etal~\cite{cao2021videosr} propose to use self attention among local patches within a video.

\begin{figure*}[!tbp]
\captionsetup{font=small}%
\scriptsize
\begin{center}
\hspace{-0.3cm}
\begin{overpic}[width=17.9cm]{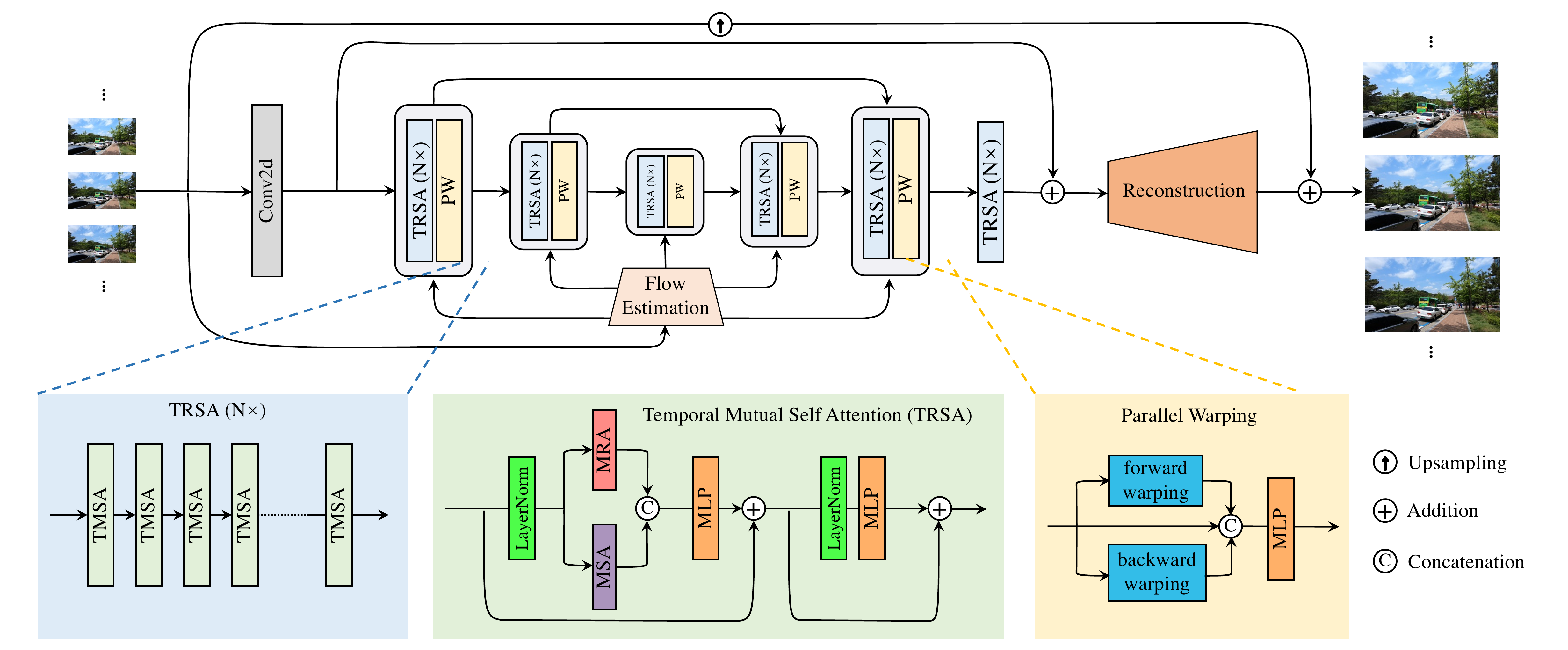}
\end{overpic}
\end{center}\vspace{-0.5cm}
\caption{The framework of the proposed Video Restoration Transformer (VRT). Given $T$ low-quality input frames, VRT reconstructs $T$ high-quality frames in parallel. It jointly extracts features, deals with misalignment, and fuses temporal information at multiple scales. On each scale, it has two kinds of modules: temporal mutual self attention (TMSA, see Sec.~\ref{sec:mama}) and parallel warping (see Sec.~\ref{sec:paw}). The downsampling and upsampling operations between different scales are omitted for clarity.}
\label{fig:framework}
\end{figure*}

\subsection{Vision Transformer}
Recently, Transformer-based models~\cite{vaswani2017transformer, shazeer2020geglu, li2021trear, wick2021transformer} have achieved promising performance in various vision tasks, such as image recognition~\cite{dosovitskiy2020ViT, carion2020DETR, li2021localvit, wick2021transformer, liu2021swin, guo2021deep, sun2021boosting, liu2021swin, liu2021transformer, liu2020deep} and image restoration~\cite{chen2021IPT, wang2021uformer, liang21swinir}. Some methods have tried to use Transformer for video modelling by extending the attention mechanism to the temporal dimension~\cite{bertasius2021timesformer, arnab2021vivit, neimark2021videotransfomernetwork, liu2021videoSwinT, li2021trear}. However, most of them are designed for visual recognition, which are fundamentally different from restoration tasks. They are more focused on feature fusion than on alignment. Cao~\etal~\cite{cao2021videosr} propose a CNN-transformer hybrid network for video super-resolution (SR) based on spatial-temporal convolutional self attention. However, it does not make full use of local information within each patch and suffers from border artifacts during testing.

\section{Video Restoration Transformer}
\subsection{Overall Framework}
\label{sec:overall}
Let $I^{\textit{LQ}}\in \mathbb{R}^{T\times H\times W\times C_{in}}$ be a sequence of low-quality (LQ) input frames and $I^{\textit{HQ}}\in \mathbb{R}^{T\times sH\times sW\times C_{out}}$ be a sequence of high-quality (HQ) target frames. $T$, $H$, $W$, $C_{in}$ and $C_{out}$ are the frame number, height, width and input channel number and output channel number, respectively. $s$ is the upscaling factor, which is larger than 1 (\eg, for video SR) or equal to 1 (\eg, for video deblurring). The proposed Video Restoration Transformer (VRT) aims to restore $T$ HQ frames from $T$ LQ frames in parallel for various video restoration tasks, including video SR, deblurring, denoising, \etc. As illustrated in Fig.~\ref{fig:framework}, VRT can be divided into two parts: feature extraction and reconstruction.

\vspace{-0.4cm}
\paragraph{Feature extraction.}
At the beginning, we extract shallow features $I^{\textit{SF}}\in \mathbb{R}^{T\times H\times W\times C}$ by a single spatial 2D convolution from the LQ sequence $I^{\textit{LQ}}$. After that, based on \cite{ronneberger2015u}, we propose a multi-scale network that aligns frames at different image resolutions. More specifically, when the total scale number is $S$, we downsample the feature for $S-1$ times by squeezing each $2\times 2$ neighborhood to the channel dimension and reducing the channel number to the original number via a linear layer. Then, we upsample the feature gradually by unsqueezing the feature back to its original size. In such a way, we can extract features and deal with object or camera motions at different scales by two kinds of modules: temporal mutual self attention (TMSA, see~\ref{sec:mama}) and parallel warping (see~\ref{sec:paw}). Skip connections are added for features of same scales. Finally, after multi-scale feature extraction, alignment and fusion, we add several TMSA modules for further feature refinement and obtain the deep feature $I^{\textit{DF}}\in \mathbb{R}^{T\times H\times W\times C}$.

\begin{figure*}
 \begin{subfigure}{.41\textwidth}
   \centering
   \begin{overpic}[width=1\textwidth]{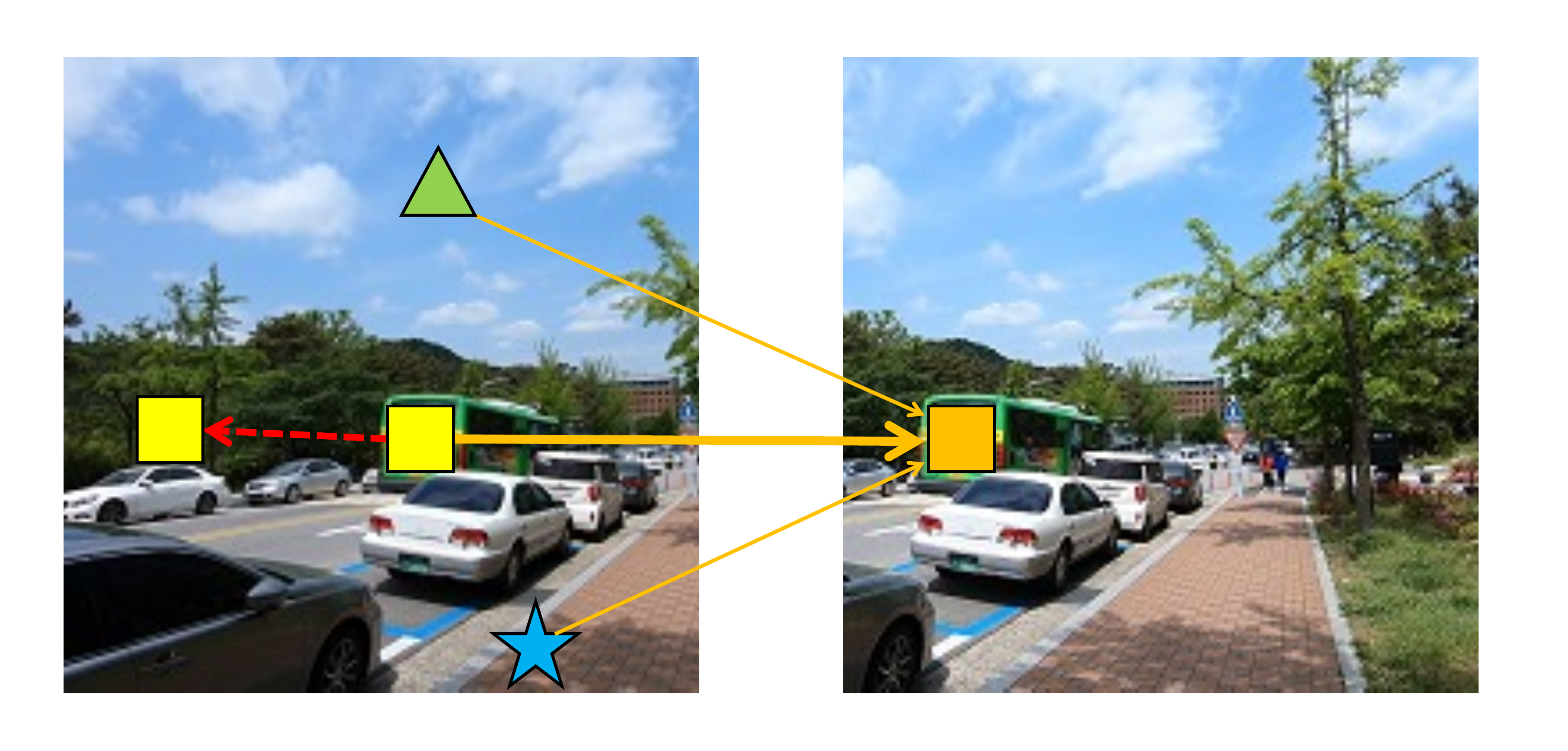}
\put(10,-2){\color{black}{\scriptsize supporting frame $X^S$ }}
\put(61,-2){\color{black}{\scriptsize reference frame $X^R$}}
\put(45,28.5){\color{black}{\scriptsize \color{black}{$A_{i,m}$}}}
\put(45,22){\color{black}{\scriptsize \color{black}{$A_{i,k}$}}}
\put(45,9){\color{black}{\scriptsize \color{black}{$A_{i,n}$}}}
     \end{overpic}
    \vspace{0.23cm}
   \caption{Mutual attention}
   \label{fig:mama_warp}
 \end{subfigure}\hspace{1cm}
 \begin{subfigure}{.5\textwidth}
   \centering
      \begin{overpic}[width=1\textwidth]{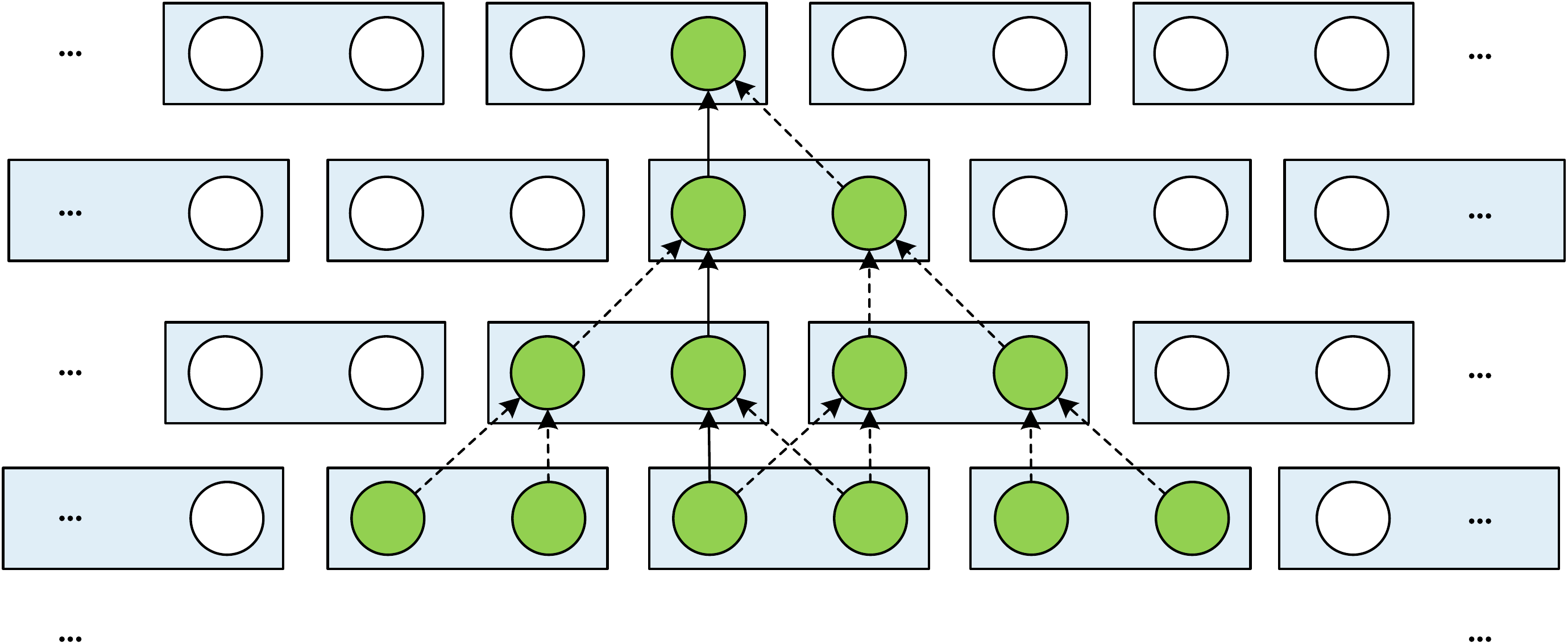}
\put(-10.3,42){\color{black}{\scriptsize (layer)}}
\put(-10,37){\color{black}{\scriptsize $i+3$}}
\put(-10,27){\color{black}{\scriptsize $i+2$}}
\put(-10,17){\color{black}{\scriptsize $i+1$}}
\put(-8,7){\color{black}{\scriptsize $i$}}
\put(11,-0.75){\color{black}{\scriptsize $t-3$}}
\put(21,-0.75){\color{black}{\scriptsize $t-2$}}
\put(33,-0.75){\color{black}{\scriptsize $t-1$}}
\put(45,-0.75){\color{black}{\scriptsize $t$}}
\put(52,-0.75){\color{black}{\scriptsize $t+1$}}
\put(63,-0.75){\color{black}{\scriptsize $t+2$}}
\put(73,-0.75){\color{black}{\scriptsize $t+3$}}
\put(84,-0.75){\color{black}{\scriptsize $t+4$}}
\put(98,-0.75){\color{black}{\scriptsize (frame)}}
     \end{overpic}
     \vspace{0.03cm}
   \caption{Stacking of temporal mutual self attention (TMSA)}
   \label{fig:mama_temporal}
 \end{subfigure}
 \caption{Illustrations for mutual attention and temporal mutual self attention (TMSA). In \ref{fig:mama_warp}, we let the orange square (the $i$-th element of the reference frame) query elements in the supporting frame and use their weighted features as a new representation for the orange square. The weights are shown around solid arrows (we only show three examples for clarity). When $A_{i,k}\rightarrow 1$ and the rest $A_{i,j}\rightarrow 0 (j\neq k)$, the mutual attention equals to warping the yellow square to the position of the orange square (illustrated as a dashed arrow). \ref{fig:mama_temporal} shows a stack of temporal mutual self attention (TMSA) layers. The sequence is partitioned into 2-frame clips at each layer and shifted for every other layer to enable cross-clip interactions. Dashed lines represent information fusion among different frames.}
 \label{fig:mama}
\end{figure*}
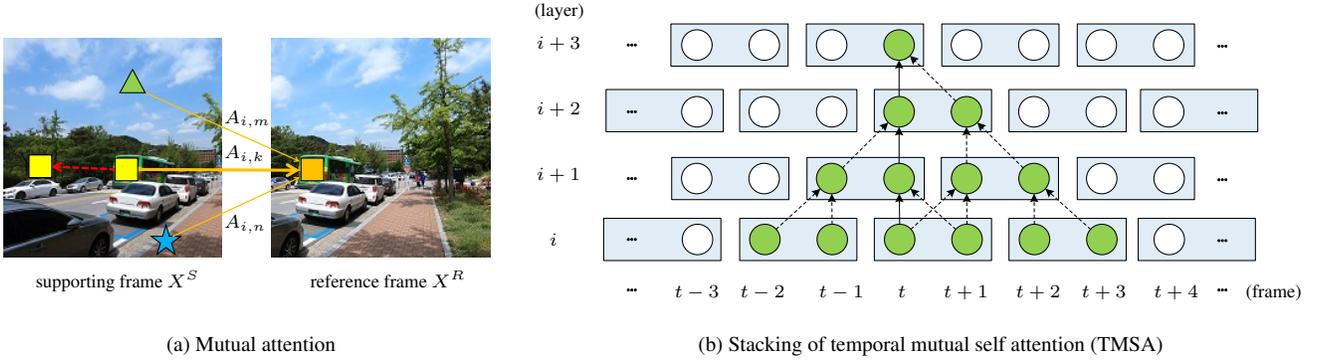

\vspace{-0.4cm}
\paragraph{Reconstruction.}
After feature extraction, we reconstruct the HQ frames from the addition of shallow feature $I^{\textit{SF}}$ and deep feature $I^{\textit{DF}}$. Different frames are reconstructed independently based on their corresponding features. Besides, to ease the burden of feature learning, we employ global residual learning and only predict the residual between the bilinearly upsampled LQ sequence and the ground-truth HQ sequence. In practice, different reconstruction modules are used for different restoration tasks. For video SR, we use the sub-pixel convolution layer~\cite{shi2016subpixel} to upsample the feature by a scale factor of $s$. For video deblurring, a single convolution layer is enough for reconstruction. Apart from this, the architecture designs are kept the same for all tasks.

\vspace{-0.4cm}
\paragraph{Loss function.}
For fair comparison with existing methods, we use the commonly used Charbonnier loss~\cite{charbonnier1994Charbonnier} between the reconstructed HQ sequence $I^{\textit{RHQ}}$ and the ground-truth HQ sequence $I^{\textit{HQ}}$ as
\begin{equation}
\mathcal{L}= \sqrt{\|I^{\textit{RHQ}}-I^{\textit{HQ}}\|^2+\epsilon^2},
\end{equation}
where $\epsilon$ is a constant that is empirically set as $10^{-3}$. %

\subsection{Temporal Mutual Self Attention}
\label{sec:mama}
In this section, based on the attention mechanism~\cite{vaswani2017transformer, li2021trear, wick2021transformer}, we first introduce the mutual attention and then propose the temporal mutual self attention (TMSA).

\vspace{-0.4cm}
\paragraph{Mutual attention.}
Given a reference frame feature $X^R\in \mathbb{R}^{N\times C}$ and a supporting frame feature $X^S\in \mathbb{R}^{N\times C}$, where $N$ is the number of feature elements and $C$ is the channel number, we compute the \textit{query} $Q^R$, \textit{key} $K^S$ and \textit{value} $V^S$ from $X^R$ and $X^S$ by linear projections as 
\begin{equation}
Q^R=X^R P^Q,\quad K^S=X^S P^K,\quad V^S=X^S P^V,
\end{equation}
where $P^Q, P^K, P^V\in \mathbb{R}^{C\times D}$ are projection matrices. $D$ is the channel number of projected features. Then, we use $Q^R$ to query $K^S$ in order to generate the attention map $A=\text{SoftMax}(Q^R (K^S)^T/\sqrt{D})\in \mathbb{R}^{N\times N}$, which is then used for weighted sum of $V^S$. This is formulated as
\begin{equation}
\text{MA}(Q^R,K^S,V^S)=\text{SoftMax}(Q^R (K^S)^T/\sqrt{D})V^S,
\label{eq:ma}
\end{equation}
where $\text{SoftMax}$ means the row softmax operation.

Since $Q^R$ and $K^S$ come from $X^R$ and $X^S$, respectively, $A$ reflects the correlation between elements in the reference image and the supporting image. For clarity, we rewrite Eq.~\eqref{eq:ma} for the $i$-th element of the reference image as
\begin{equation}
Y_{i,:}^{R}=\sum_{j=1}^N A_{i,j}V_{j,:}^S,
\label{eq:mama_weighted}
\end{equation}
where $Y_{i,:}^{R}$ refers to the new feature of the $i$-th element in the reference frame. As shown in Fig.~\ref{fig:mama_warp}, when $K_{k,:}^S$ (\eg, the yellow square from the supporting frame) is the most similar element to $Q_{i,:}^R$ (\eg, the orange square from the reference frame), $A_{i,k}>A_{i,j}$ holds for all $j\neq k$ ($j\leq N$). When all $K_{j,:}^S (j\neq k)$ are very dissimilar to $Q_i^R$, we have 
\begin{equation}
\begin{cases}
A_{i,k}\rightarrow 1,\\
A_{i,j}\rightarrow 0, ~~~~~~~~~~~~~~~~~~~for~~ j\neq k, j\leq N.
\label{eq:mama_extreme}
\end{cases}
\end{equation}

In this extreme case, by combining Eq.~\eqref{eq:mama_weighted} and~\eqref{eq:mama_extreme}, we have $Y_{i,:}^{R}=V_{k,:}^S$, which moves the $k$-th element in the supporting frame to the position of the $i$-th element in the reference frame (see the dashed red line in Fig.~\ref{fig:mama_warp}). This equals to image warping given an optical flow vector. When $A_{i,k}\rightarrow 1$ does not hold, Eq.~\eqref{eq:mama_weighted} can be regarded as a ``soft'' version of image warping. In practice, the reference frame and supporting frame can be exchanged, allowing mutual alignment between two frames. Besides, similar to multi-head self attention, we can also perform the attention for $h$ times and concatenate the results as multi-head mutual attention (MMA).

Particularly, mutual attention has several benefits over the combination of explicit motion estimation and image warping. First, mutual attention can adaptively preserve information from the supporting frame than image warping, which only focuses on the target pixel. It also avoids black hole artifacts when there is no matched positions. Second, mutual attention does not have the inductive biases of locality, which is inherent to most CNN-based motion estimation methods~\cite{dosovitskiy2015flownet, ranjan2017spynet,pytorch-spynet, sun2018pwc} and may lead to performance drop when two neighboring objects move towards different directions. Third, mutual attention equals to conducting motion estimation and warping on image features in a joint way. In contrast, optical flows are often estimated on the input RGB image and then used for warping on features~\cite{chan2021basicvsr, chan2021basicvsr++}. Besides, flow estimation on RGB images is often not robust to lighting variation, occlusion and blur~\cite{xue2019TOFlow-Vimeo-90K}.

\vspace{-0.4cm}
\paragraph{Temporal mutual self attention (TMSA).}
Mutual attention is proposed for joint feature alignment between two frames. To extract and preserve feature from the current frame, we use mutual attention together with self attention. Let $X\in\mathbb{R}^{2\times N\times C}$ represent two frames, which can be split into $X_1\in\mathbb{R}^{1\times N\times C}$ and $X_2\in\mathbb{R}^{1\times N\times C}$. We use multi-head mutual attention (MMA) on $X_1$ and $X_2$ for two times: warping $X_1$ towards $X_2$ and warping $X_2$ towards $X_1$. The warped features are combined and then concatenated with the result of multi-head self attention (MSA), followed by a multi-layer perceptron (MLP) for the purpose of dimension reduction. After that, another MLP is added for further feature transformation. Two LayerNorm (LN) layers and two residual connections are also used as shown in the green box of Fig.~\ref{fig:framework}. The whole process formulated as follows
\begin{equation}
\begin{split}
&X_1, X_2 = \text{Split}_0(\text{LN}(X))\\
&Y_1, Y_2=\text{MMA}(X_1, X_2),\text{MMA}(X_2, X_1)\\
&Y_3=\text{MSA}([X_1, X_2])\\
&X=\text{MLP}(\text{Concat}_2(\text{Concat}_0(Y_1, Y_2), Y_3))+X\\
&X=\text{MLP}(\text{LN}(X))+X
\end{split}
\label{eq:masa}
\end{equation}
where the subscripts of $\text{Split}$ and $\text{Concat}$ refer to the specified dimensions. However, due to the design of mutual attention, Eq.~\eqref{eq:masa} can only deal with two frames at a time.

One naive way to extend Eq.~\eqref{eq:masa} for $T$ frames is to deal with frame-to-frame pairs exhaustively, resulting in the computational complexity of $\mathcal{O}(T^2)$. Inspired by the shifted window mechanism~\cite{liu2021swin, liu2021videoSwinT}, we propose the temporal mutual self attention (TMSA) to remedy the problem. TMSA first partitions the video sequence into non-overlapping 2-frame clips and then applies Eq.~\eqref{eq:masa} to them in parallel. Next, as shown in Fig.~\ref{fig:mama_temporal}, it shifts the sequence temporally by 1 frame for every other layer to enable cross-clip connections, reducing the computational complexity to $\mathcal{O}(T)$. The temporal receptive field size is increased when multiple TMSA modules are stacked together. Specifically, at layer $i$ ($i\geq 2$), one frame can utilize information from up to $2(i-1)$ frames.

\vspace{-0.4cm}
\paragraph{Discussion.}
Video restoration tasks often need to process high-resolution frames. Since the complexity of attention is quadratic to the number of elements within the attention window, global attention on the full image is often impractical. Therefore, following~\cite{liu2021swin, liang21swinir}, we partition each frame spatially into non-overlapping $M\times M$ local windows, resulting in $\frac{HW}{M^2}$ windows. Shifted window mechanism (with the shift of $\lfloor\frac{M}{2}\rfloor \times \lfloor\frac{M}{2}\rfloor$ pixels) is also used spatially to enable cross-window connections. Besides, although stacking multiple TMSA modules allows for long-distance temporal modelling, distant frames are not directly connected. As will show in the ablation study, using only a small temporal window size cannot fully exploit the potential of the model. Therefore, we use larger temporal window size for the last quarter of TMSA modules to enable direct interactions between distant frames.

\subsection{Parallel Warping}
\label{sec:paw}
Due to spatial window partitioning, the mutual attention mechanism may not be able to deal with large motions well. Hence, as shown in the orange box of Fig.~\ref{fig:framework}, we use feature warping at the end of each network stage to handle large motions. For any frame feature $X_t$, we calculate the optical flows of its neighbouring frame features $X_{t-1}$ and $X_{t+1}$, and warp them towards the frame $X_t$ as $\hat{X}_{t-1}$ and $\hat{X}_{t+1}$ (\ie, backward and forward warping). Then, we concatenate them with the original feature and use an MLP for feature fusion and dimension reduction. Specifically, following~\cite{chan2021basicvsr++}, we predict the residual flow by a flow estimation model and use deformable convolution~\cite{dai2017deformable} for deformable alignment. More details are provided in the supplementary.

\section{Experiments}
\subsection{Experimental Setup}
For video SR, we use 4 scales for VRT. On each scale, we stack 8 TMSA modules, the last two of which use a temporal window size of 8. The spatial window size $M\times M$, head size $h$, and channel size $C$ are set to $8\times 8$,  6 and 120, respectively. After 7 multi-scale feature extraction stages, we add 24 TMSA modules (only with self attention) for further feature extraction before reconstruction. More details are provided in the supplementary.

\begin{table*}[!t]
    \caption{Quantitative comparison (average PSNR/SSIM) with state-of-the-art methods for \textbf{video super-resolution ($\times 4$)} on \textbf{REDS4}~\cite{nah2019ntireREDS}, \textbf{Vimeo-90K-T}~\cite{xue2019TOFlow-Vimeo-90K}, \textbf{Vid4}~\cite{liu2013bayesianVid4} and \textbf{UDM10}~\cite{yi2019pfnl_udm}. Best and second best results are in \R{red} and \B{blue} colors, respectively. $^\dagger$We currently do not have enough GPU memory to train the fully parallel model VRT on 30 frames.}%
    \label{tab:vsr_quan}
    \vspace{-0.6cm}
    \begin{center}\scalebox{0.84}{
            \tabcolsep=0.1cm
            \begin{tabular}{|l|c|c|c||c||c|c||c|c|c|}
                \hline
                \multirow{2}{*}{\makecell{\vspace{-0.5cm}Method}}                  &   \multirow{2}{*}{\scalebox{0.7}{\makecell{Training\\Frames\\(REDS/\\Vimeo-90K)}}}        &    \multirow{2}{*}{\vspace{-0.5cm}\makecell{Params\\(M)}}          &    \multirow{2}{*}{\vspace{-0.5cm}\makecell{Runtime\\(ms)}}       & \multicolumn{3}{c||}{BI degradation} & \multicolumn{3}{c|}{BD degradation}                     
                
                \\ \cline{5-10}                       &  &  &  & \makecell{REDS4~\cite{nah2019ntireREDS}\\(RGB channel)}           & \makecell{\scalebox{0.7}{Vimeo-90K-T}~\cite{xue2019TOFlow-Vimeo-90K}\\(Y channel)}  & \makecell{Vid4~\cite{liu2013bayesianVid4}\\(Y channel)} & \makecell{UDM10~\cite{yi2019pfnl_udm}\\(Y channel)} & \makecell{\scalebox{0.7}{Vimeo-90K-T}~\cite{xue2019TOFlow-Vimeo-90K}\\(Y channel)} & \makecell{Vid4~\cite{liu2013bayesianVid4}\\(Y channel)} \\ \hline\hline
                Bicubic                             & -        & -          & -            & 26.14/0.7292                         & 31.32/0.8684                       & 23.78/0.6347                & 28.47/0.8253                   & 31.30/0.8687                    & 21.80/0.5246                \\
                SwinIR~\cite{liang21swinir}  & -             & 11.9          & -            &         29.05/0.8269                &           35.67/0.9287             &       25.68/0.7491          &      35.42/0.9380              &     34.12/0.9167                &        25.25/0.7262         \\ 
                SwinIR-ft~\cite{liang21swinir}  & 1/1             & 11.9          & -            &         29.24/0.8319               &           35.89/0.9301             &      25.69/0.7488           &          36.76/0.9467          &        35.70/0.9293             &        25.62/0.7498         \\ \hline
                TOFlow~\cite{xue2019TOFlow-Vimeo-90K}   & 5/7                & -          & -            & 27.98/0.7990                         & 33.08/0.9054                       & 25.89/0.7651                & 36.26/0.9438                   & 34.62/0.9212                    & 25.85/0.7659                           \\
                FRVSR~\cite{sajjadi2018FRVSR}   & 10/7             & 5.1        & 137          & -                                    & -                                  & -                           & 37.09/0.9522                   & 35.64/0.9319                    & 26.69/0.8103                \\
                DUF~\cite{jo2018DUF}      & 7/7                  & 5.8        & 974          & 28.63/0.8251                         & -                                  & 27.33/0.8319                           & 38.48/0.9605                   & 36.87/0.9447                    & 27.38/0.8329                \\
                PFNL~\cite{yi2019pfnl_udm}  & 7/7              & 3.0        & 295          & 29.63/0.8502                         & 36.14/0.9363                       & 26.73/0.8029                & 38.74/0.9627                   & -                               & 27.16/0.8355                \\
                RBPN~\cite{haris2019recurrent}  & 7/7             & 12.2       & 1507         & 30.09/0.8590                         & 37.07/0.9435                       & 27.12/0.8180                & 38.66/0.9596                   & 37.20/0.9458                    & 27.17/0.8205                           \\
                
                MuCAN~\cite{li2020mucan}   & 5/7                 & -          & -            & 30.88/0.8750                         & 37.32/0.9465                       & -                           & -                              & -                               & -                           \\
                RLSP~\cite{fuoli2019rlsp}  & -/7             & 4.2        & 49           & -                                    & -                                  & -                           & 38.48/0.9606                   & 36.49/0.9403                    & 27.48/0.8388                \\
                TGA~\cite{isobe2020tga}  & -/7                  & 5.8        & 384            & -                                    & -                                  & -                           & 38.74/0.9627                              & 37.59/0.9516                    & 27.63/0.8423                \\
                RSDN~\cite{isobe2020rsdn}  & -/7                & 6.2        & 94           & -                                    & -                                  & -                           & 39.35/0.9653                   & 37.23/0.9471                    & 27.92/0.8505                \\
                 RRN~\cite{isobe2020rrn}  & -/7             & 3.4        & 45           & -                                    & -                                  & -                           & 38.96/0.9644                   & -                               & 27.69/0.8488                \\
                FDAN~\cite{lin2021fdan}   & -/7               & 9.0       & -           & -                                    & -                                  & -                           & 39.91/0.9686                  & 37.75/0.9522                   & 27.88/0.8508                \\
               EDVR~\cite{wang2019edvr}  & 5/7                  & 20.6       & 378          & 31.09/0.8800                         & {37.61}/{0.9489}             & 27.35/0.8264                & 39.89/0.9686                   & 37.81/0.9523                    & 27.85/0.8503                \\
                 GOVSR~\cite{yi2021omniscient}   & -/7               & 7.1        & 81           & -                                    & -                                  &      -                    &       40.14/0.9713               &          37.63/0.9503           & 28.41/0.8724               \\
                VSRT~\cite{cao2021videosr}  & 5/7                   & 32.6       & -          & \B{31.19/0.8815}                         & 37.71/0.9494             & 27.36/0.8258                & -                   & -                    & -               \\
                \textbf{VRT} (ours)                    &    6/-     &      30.7      &          236      &    \R{31.60/0.8888}          &    -   &      -    &     -      &     -  &  -\\ %
                \hline
                \mbox{BasicVSR}~\cite{chan2021basicvsr}  & 15/14   & 6.3        & 63           & 31.42/0.8909                         & 37.18/0.9450                       & 27.24/0.8251                & 39.96/{0.9694}              & 37.53/0.9498                    & 27.96/0.8553                \\
                \mbox{IconVSR}~\cite{chan2021basicvsr}  & 15/14     & 8.7        & 70           & \B{{31.67}/{0.8948}}               & 37.47/0.9476                       & {27.39}/{0.8279}      & {40.03}/{0.9694}         & {37.84}/{0.9524}          & {28.04}/{0.8570}      \\ 
                
                {BasicVSR++}\cite{chan2021basicvsr++}  & 30/14                        & 7.3        & 77           & ~{\color{gray}{{32.39}/{0.9069}}}\color{black}{$^\dagger$}               & \B{{37.79}/{0.9500}}        & \B{{27.79}/{0.8400}}      & \B{{40.72}/{0.9722}}         & \B{{38.21}/{0.9550}}          & \B{{29.04}/{0.8753}}      \\ 
                \textbf{VRT} (ours)                &    16/7     &     35.6       &    243            &     \R{32.19/0.9006}         &   \R{38.20/0.9530}    &    \R{27.93/0.8425}      &    \R{41.05/0.9737}       &   \R{38.72/0.9584}        &  \R{29.42/0.8795}   \\
                \hline
            \end{tabular}}
        \vspace{-0.4cm}
    \end{center}
\end{table*}

\begin{figure*}[!htbp]
	\captionsetup{font=small}
	
	\centering
	\scriptsize
	
	\renewcommand{\h}{0.105}
	\renewcommand{\wa}{0.12}
	\newcommand{\wb}{0.16}
	\renewcommand{\g}{-0.7mm}
	\renewcommand{\tabcolsep}{1.8pt}
	\renewcommand{\arraystretch}{1}
	\resizebox{1.00\linewidth}{!} {
		\begin{tabular}{cc}
			
			\renewcommand{\name}{figures/sr/00000021_}
			\renewcommand{\h}{0.12}
			\renewcommand{\w}{0.2}
			\begin{tabular}{cc}
				\begin{adjustbox}{valign=t}
					\begin{tabular}{c}%
		         	\includegraphics[trim={84 0 0 0 },clip, width=0.354\textwidth]{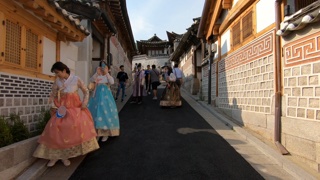}
						\\
						Frame 021, Clip 011, REDS~\cite{nah2019ntireREDS} 
					\end{tabular}
				\end{adjustbox}
				\begin{adjustbox}{valign=t}
					\begin{tabular}{cccccc}
						\includegraphics[trim={237 152 65 15 },clip,height=\h \textwidth, width=\w \textwidth]{\name LQ.jpeg} \hspace{\g} &
						\includegraphics[trim={950 610 260 60},clip,height=\h \textwidth, width=\w \textwidth]{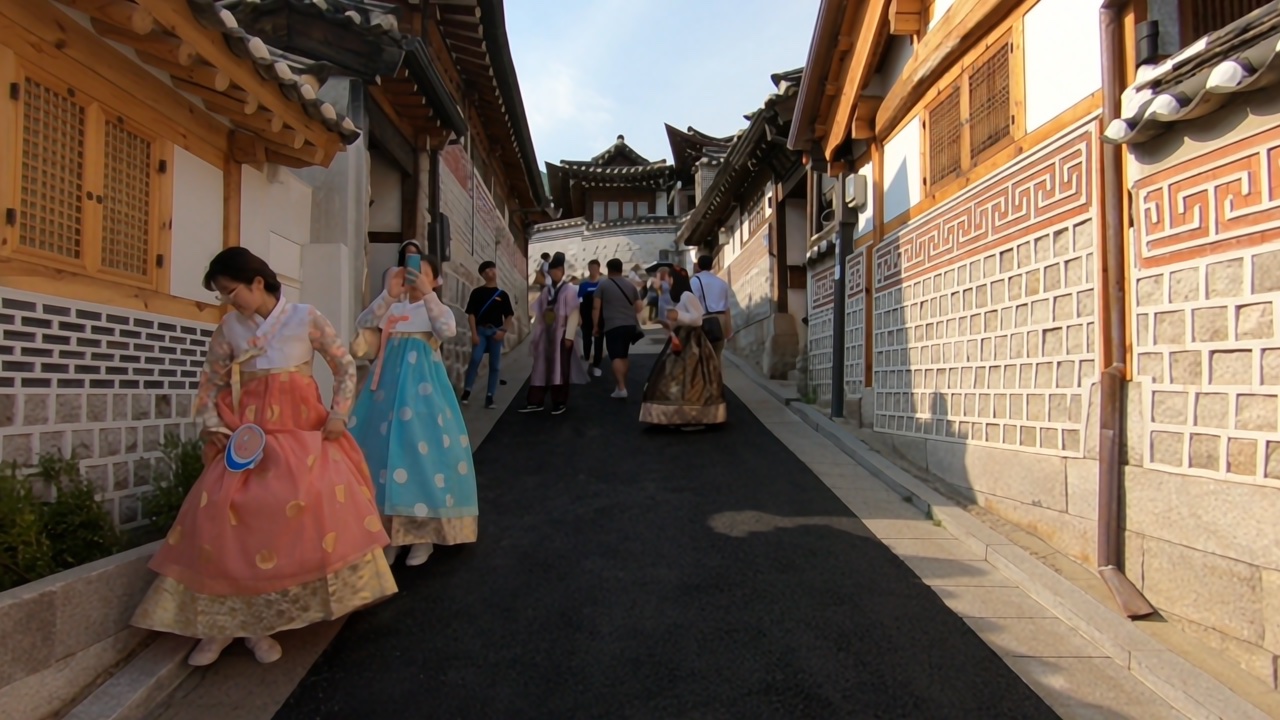} \hspace{\g} &
						\includegraphics[trim={950 610 260 60},clip,height=\h \textwidth, width=\w \textwidth]{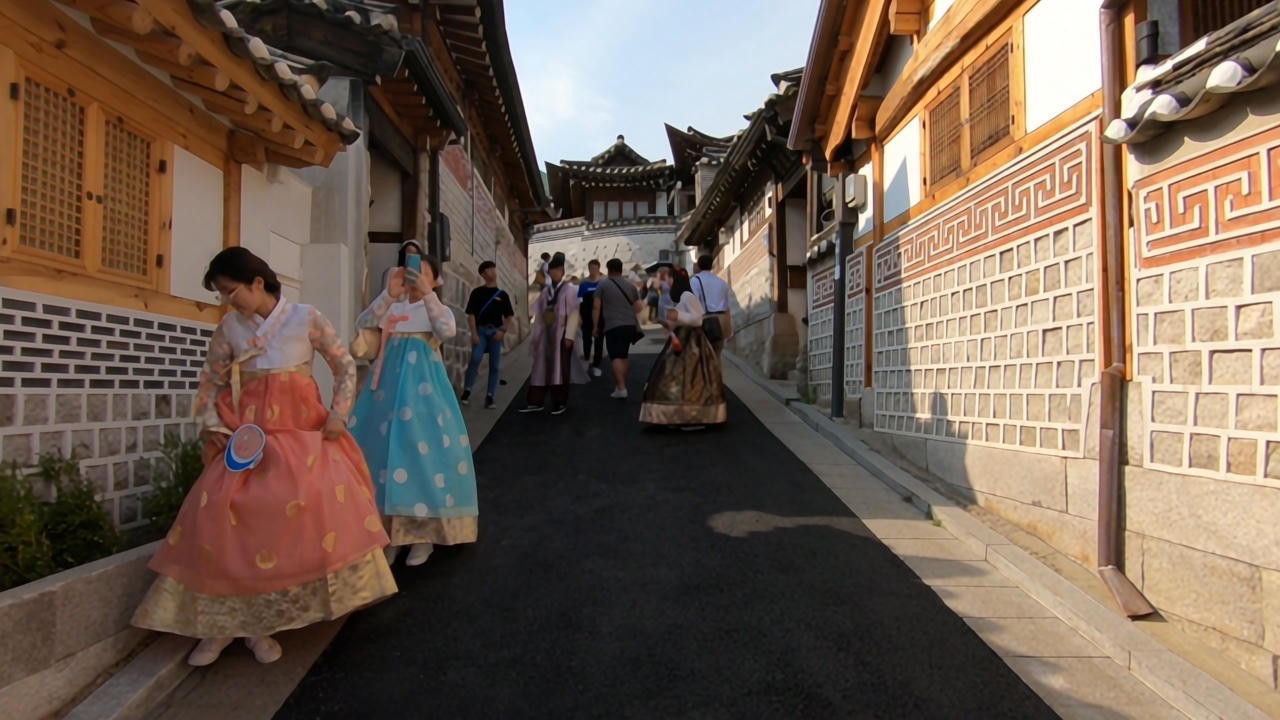} &
						\includegraphics[trim={950 610 260 60},clip,height=\h \textwidth, width=\w \textwidth]{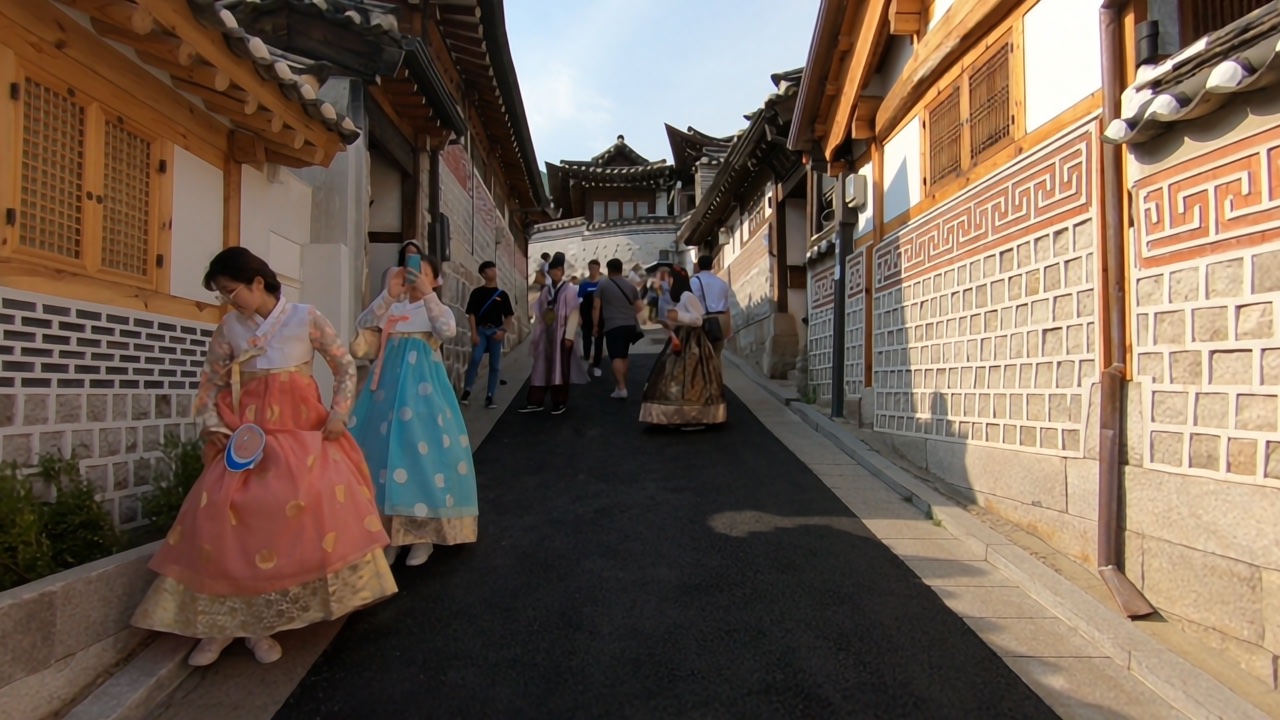} \hspace{\g} 
						\\
						LQ ($\times 4$) &
						EDVR~\cite{wang2019edvr} & VSRT~\cite{cao2021videosr} &
						BasicVSR~\cite{chan2021basicvsr} 
						\\
						\vspace{-1.5mm}
						\\
						\includegraphics[trim={950 610 260 60},clip,height=\h \textwidth, width=\w \textwidth]{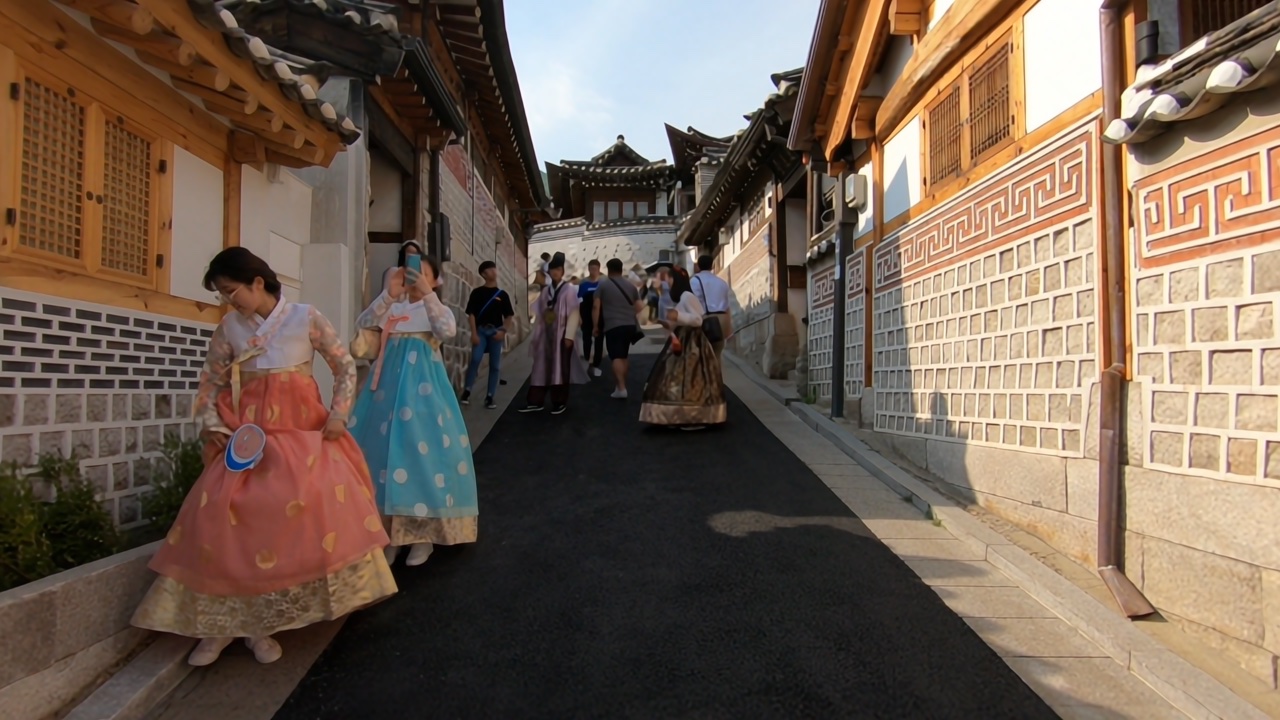} \hspace{\g} &
						\includegraphics[trim={950 610 260 60},clip,height=\h \textwidth, width=\w \textwidth]{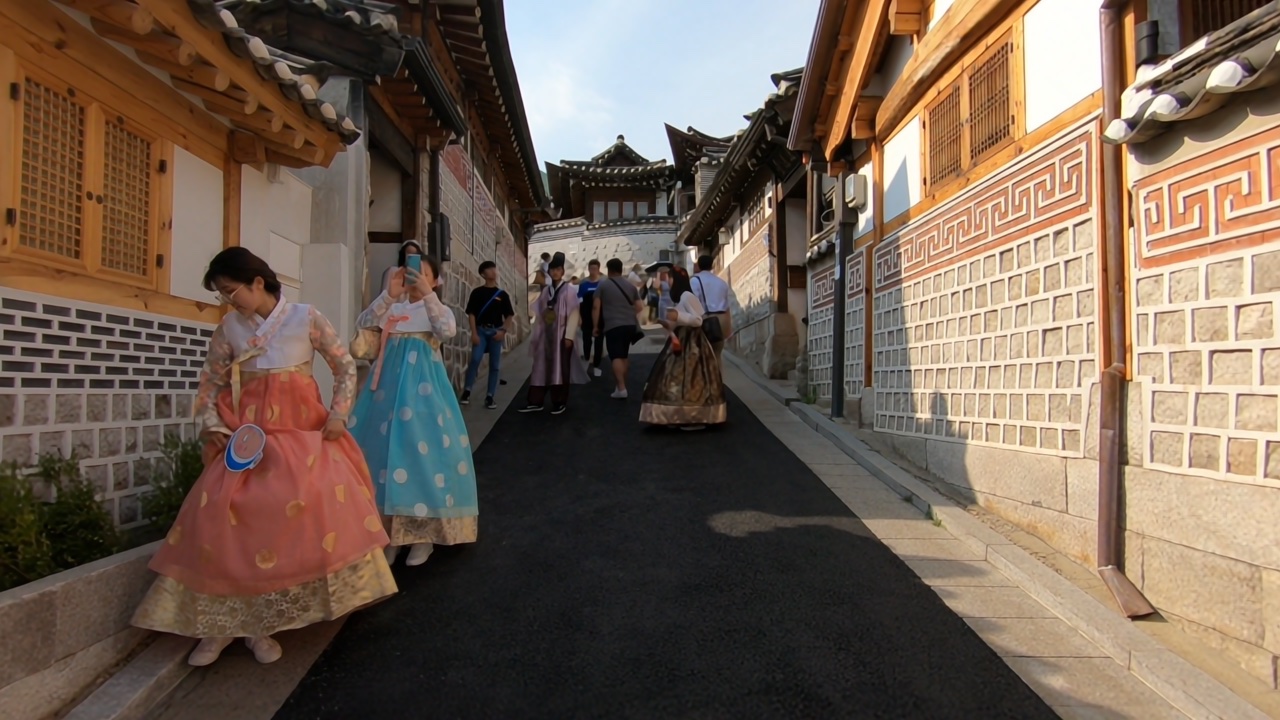} \hspace{\g} &
						\includegraphics[trim={950 610 260 60},clip,height=\h \textwidth, width=\w \textwidth]{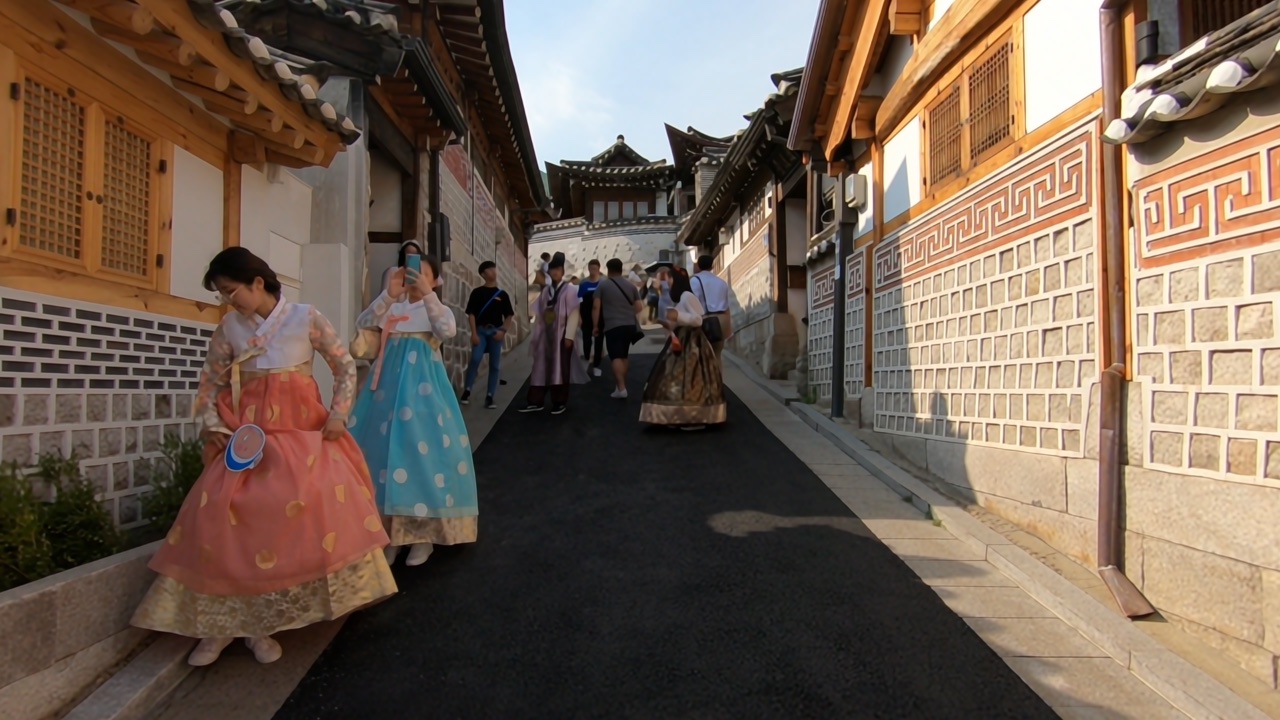}
						\hspace{\g} &		
						\includegraphics[trim={950 610 260 60 },clip,height=\h \textwidth, width=\w \textwidth]{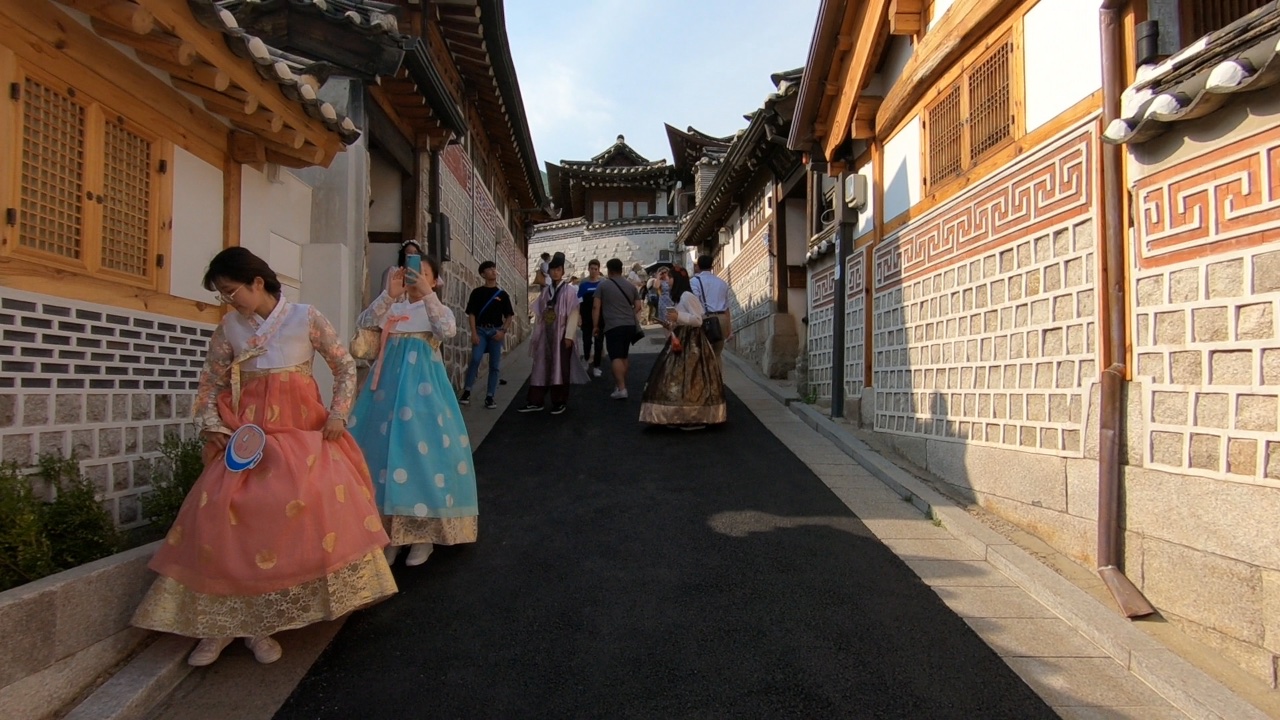} 
						\\ 
						IconVSR~\cite{chan2021basicvsr}  \hspace{\g} &	BasicVSR++~\cite{chan2021basicvsr++} \hspace{\g} &
						\textbf{VRT} (ours) &
						GT
						\\
					\end{tabular}
				\end{adjustbox}
			\end{tabular}
			
		\end{tabular}
	}
	\\ \vspace{1mm}
	\resizebox{1.00\linewidth}{!} {
	\begin{tabular}{cc}
			
			\renewcommand{\name}{figures/sr/00000022_}
			\renewcommand{\h}{0.12}
			\renewcommand{\w}{0.2}
			\begin{tabular}{cc}
				\begin{adjustbox}{valign=t}
					\begin{tabular}{c}%
		         	\includegraphics[trim={0 10 0 0 },clip, width=0.354\textwidth]{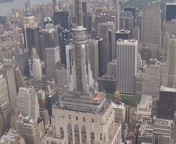}
						\\
						Frame 022, Clip city, Vid4~\cite{liu2013bayesianVid4} 
					\end{tabular}
				\end{adjustbox}
				\begin{adjustbox}{valign=t}
					\begin{tabular}{cccccc}
						\includegraphics[trim={116 85 39 37},clip,height=\h \textwidth, width=\w \textwidth]{\name LQ.jpeg} \hspace{\g} &
						\includegraphics[trim={465 340 155 150 },clip,height=\h \textwidth, width=\w \textwidth]{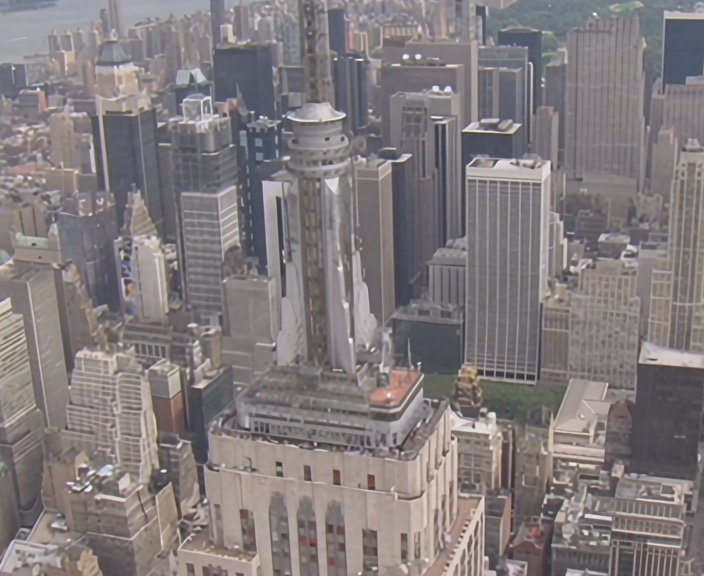} \hspace{\g} &
						\includegraphics[trim={465 340 155 150},clip,height=\h \textwidth, width=\w \textwidth]{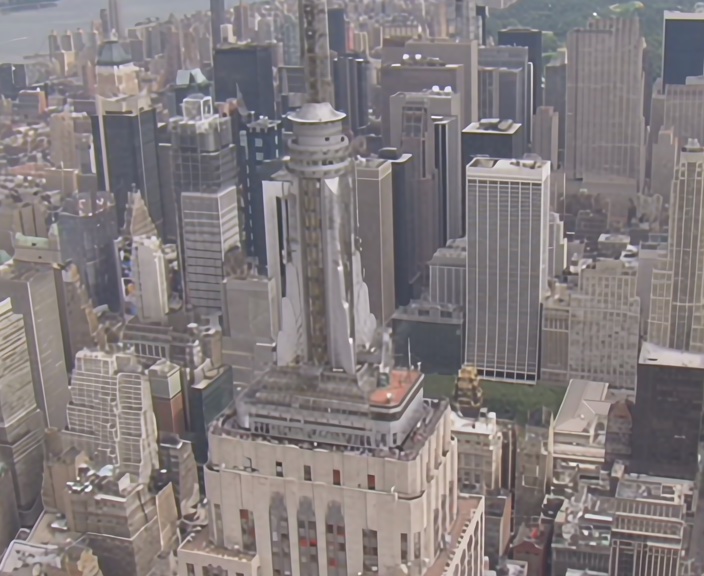} & 
						\includegraphics[trim={465 340 155 150},clip,height=\h \textwidth, width=\w \textwidth]{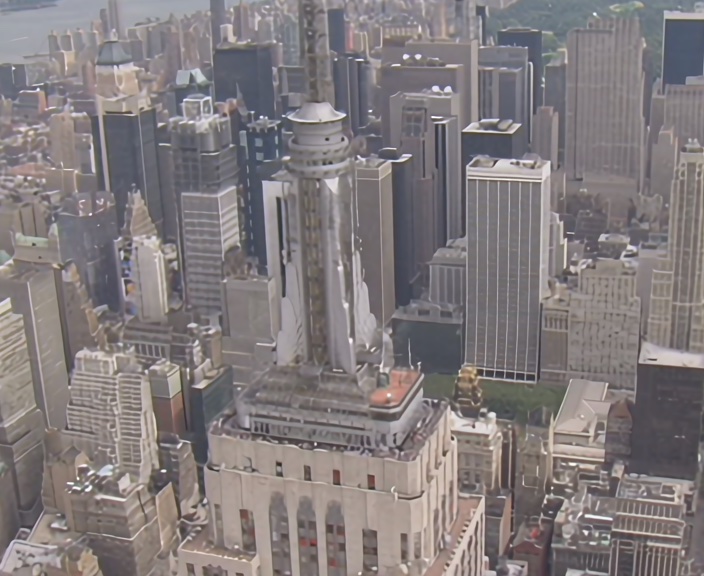} \hspace{\g} 
						\\
						LQ ($\times 4$) &
						EDVR~\cite{wang2019edvr} & VSRT~\cite{cao2021videosr} &
						BasicVSR~\cite{chan2021basicvsr} 
						\\
						\vspace{-1.5mm}
						\\
						\includegraphics[trim={465 340 155 150},clip,height=\h \textwidth, width=\w \textwidth]{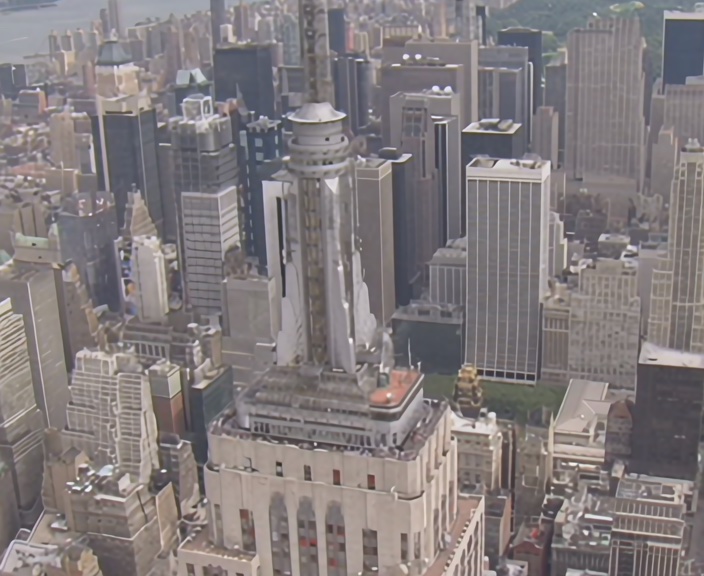} \hspace{\g} &
						\includegraphics[trim={465 340 155 150},clip,height=\h \textwidth, width=\w \textwidth]{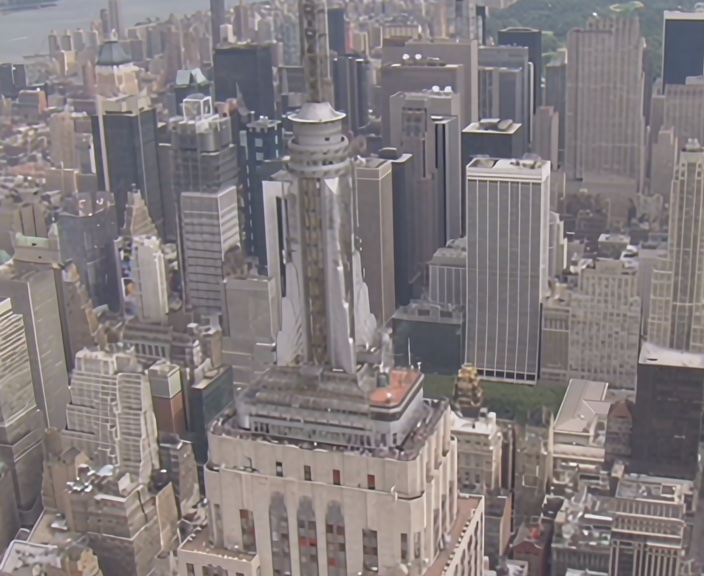} \hspace{\g} &
						\includegraphics[trim={465 340 155 150 },clip,height=\h \textwidth, width=\w \textwidth]{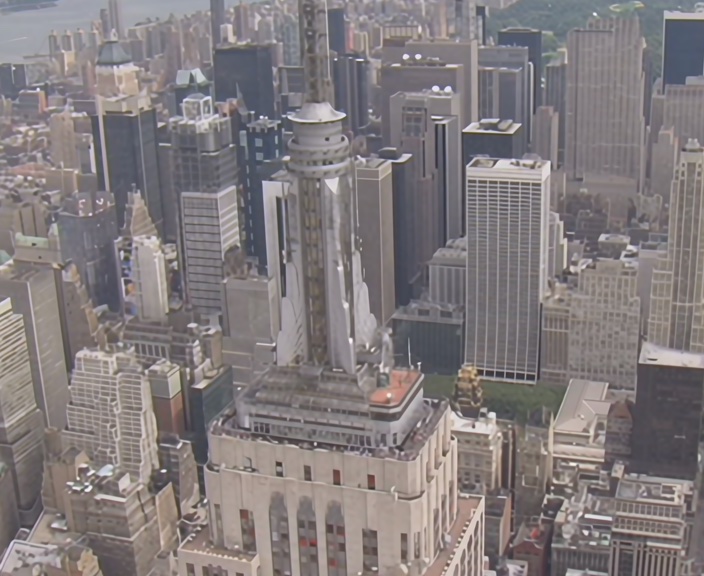}
						\hspace{\g} &		
						\includegraphics[trim={465 340 155 150},clip,height=\h \textwidth, width=\w \textwidth]{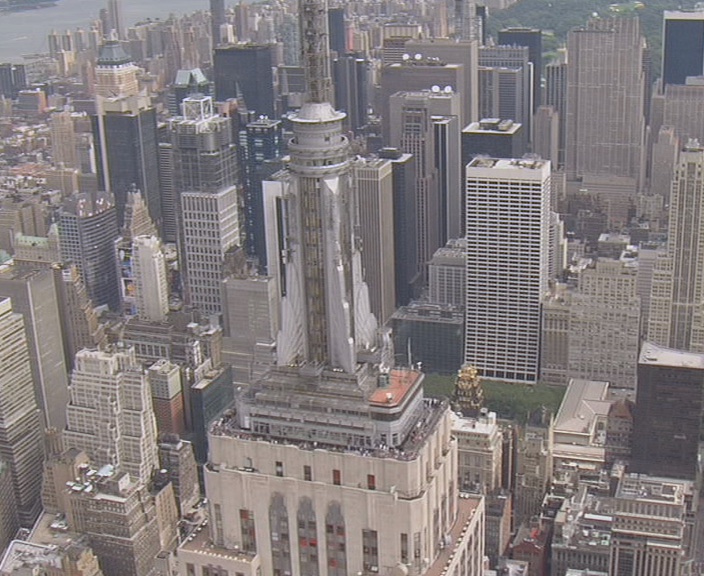} 
						\\ 
						IconVSR~\cite{chan2021basicvsr}  \hspace{\g} &	BasicVSR++~\cite{chan2021basicvsr++} \hspace{\g} &
						\textbf{VRT} (ours) &
						GT
						\\
					\end{tabular}
				\end{adjustbox}
			\end{tabular}
			
		\end{tabular}
	}
	
	\vspace{-2mm}
	\caption{Visual comparison of \textbf{{video super-resolution}} ($\times 4$) methods.} %
	  \vspace{-2mm}
	\label{fig:sr_quali}
\end{figure*}

\subsection{Video Super-Resolution}
As shown in Table~\ref{tab:vsr_quan}, we compare VRT with the state-of-the-art image and video SR methods. VRT achieves best performance for both bicubic (BI) and blur-downsamplng (BD) degradations. Specifically, when trained on the REDS~\cite{nah2019ntireREDS} dataset with short sequences, VRT outperforms VSRT by up to 0.57dB in PSNR. Compared with another representative sliding window-based model EDVR, VRT has an improvement of 0.50$\sim$1.57dB on different datasets, showing its good ability to fuse information from multiple frames. Note that VRT outputs all frames simultaneously rather than predicting them frame by frame as EDVR does. On the Vimeo-90K~\cite{xue2019TOFlow-Vimeo-90K} dataset, VRT surpasses BasicVSR++ by up to 0.38dB, although BasicVSR++ and other recurrent models may mirror the 7-frame video for training and testing. When VRT is trained on longer sequences, it shows good potential in temporal modelling and further increases the PSNR by 0.52dB. As indicated in ~\cite{cao2021videosr}, recurrent models often suffer from significant performance drops on short sequences. In contrast, VRT performs well on both short and long sequences. We note that VRT is slightly lower than the 32-frame model BasicVSR++. This is expected since VRT is only trained on 16 frames.

We also provide comparison on parameter number and runtime in Table~\ref{tab:vsr_quan}. As a parallel model, VRT needs to restore all frames at the same time, which leads to relatively larger model size and longer runtime per frame compared with recurrent models. However, VRT has the potential for distributed deployment, which is hard for recurrent models that restore a video clip recursively by design.

Visual results of different methods are shown in Fig.~\ref{fig:sr_quali}. As one can see, in accordance with its significant quantitative improvements, VRT can generate visually pleasing images with sharp edges and fine details, such as horizontal strip patterns of buildings. By contrast, its competitors suffer from either distorted textures or lost details.

\begin{table*}[!t]\scriptsize
\center
\begin{center}
\caption{Quantitative comparison (average RGB channel PSNR/SSIM) with state-of-the-art methods for \textbf{{video deblurring}} on \textbf{DVD}~\cite{su2017dvddeblur}. Following \cite{pan2020tspdeblur, li2021arvo}, all restored frames instead of randomly selected 30 frames from each test set~\cite{su2017dvddeblur} are used in evaluation. Best and second best results are in \R{red} and \B{blue} colors, respectively.}
\vspace{-3mm}
\label{tab:deblur_dvd}
\begin{tabular}{|c|c|c||c|c|c|c|c|c|c|c|c|c|c|c|}
\hline
Method & 
\makecell{\scalebox{0.9}{DeepDeblur}\\\cite{nah2017Gopro}} & %
\makecell{\scalebox{0.9}{SRN}\\\cite{tao2018scale}} & %

\makecell{\scalebox{0.9}{DBN}\\\cite{su2017dvddeblur}} &  %
\makecell{\scalebox{0.9}{DBLRNet}\\\cite{zhang2018adversarial}} & 
\makecell{STFAN\\\cite{zhou2019spatio}} & 
\makecell{\scalebox{0.9}{STTN}\\\cite{kim2018spatio}} & 
\makecell{\scalebox{0.9}{SFE}\\\cite{xiang2020deep}} &
\makecell{\scalebox{0.9}{EDVR}\\\cite{wang2019edvr}} &
\makecell{\scalebox{0.9}{TSP}\\\cite{pan2020tspdeblur}} &
\makecell{\scalebox{0.9}{PVDNet}\\\cite{son2021recurrent}} & 
\makecell{\scalebox{0.9}{GSTA}\\\cite{suin2021gated}}&
\makecell{\scalebox{0.9}{ARVo}\\\cite{li2021arvo}} &
\makecell{\textbf{\scalebox{0.9}{VRT}}\scalebox{0.9}{ (ours)}}
\\
\hline
\hline
PSNR 
& 29.85
& 30.53 %
& 30.01 
& 30.08 
& 31.24 %
& 31.61
& 31.71 
& 31.82 %
& 32.13 
& 32.31
& 32.53
& \B{32.80}
& \R{34.27 (+1.47)} 
\\
SSIM &
0.8800
& 0.8940 %
& 0.8877 
& 0.8845 
& 0.9340 %
& 0.9160
& 0.9160 
& 0.9160 %
& 0.9268 
& 0.9260
& 0.9468 %
& \B{0.9352}
& \R{0.9651 (+0.03)} %
\\
\hline             
\end{tabular}
\end{center}
\vspace{-1.2mm}
\end{table*}

\begin{table*}[!t]\scriptsize
\center
\begin{center}
\caption{Quantitative comparison (average RGB channel PSNR/SSIM) with state-of-the-art methods for \textbf{{video deblurring}} on \textbf{GoPro}~\cite{nah2017Gopro}. Best and second best results are in \R{red} and \B{blue} colors, respectively.}
\vspace{-2mm}
\label{tab:deblur_gopro}
\begin{tabular}{|c|c|c|c|c|c||c|c|c|c|c|c|c|c|}
\hline
Method & 
\makecell{\scalebox{0.9}{DeepDeblur}\\\cite{nah2017Gopro}} &
\makecell{\scalebox{0.9}{SRN}\\\cite{tao2018scale}} &
\makecell{\scalebox{0.9}{DMPHN}\\\cite{zhang2019deep}} &
\makecell{\scalebox{0.9}{SAPHN}\\\cite{suin2020spatially}} &
\makecell{\scalebox{0.9}{MPRNet}\\\cite{zamir2021MPRNet}} &
\makecell{SFE\\\cite{xiang2020deep}} & 
\makecell{\scalebox{0.9}{IFI-RNN}\\\cite{nah2019recurrent}} & 
\makecell{\scalebox{0.9}{ESTRNN}\\\cite{zhong2020efficient}} &
\makecell{\scalebox{0.9}{EDVR}\\\cite{wang2019edvr}} &
\makecell{\scalebox{0.9}{TSP}\\\cite{pan2020tspdeblur}} & 
\makecell{\scalebox{0.9}{PVDNet}\\\cite{son2021recurrent}} & 
\makecell{\scalebox{0.9}{GSTA}\\\cite{suin2021gated}}&
\makecell{\textbf{\scalebox{0.9}{VRT}}\scalebox{0.9}{ (ours)}}
\\
\hline
\hline
PSNR 
& 29.23 %
& 30.26
& 31.20
& 31.85
& \B{32.66}
& 31.01
& 31.05
& 31.07 %
& 31.54 %
& 31.67
& 31.98
& 32.10
& \R{34.81 (+2.15)}
\\
SSIM 
& 0.9162 
& 0.9342
& 0.9400
& 0.9480
& {0.9590}
& 0.9130
& 0.9110
& 0.9023 %
& 0.9260 %
& 0.9279
& 0.9280
& \B{0.9600} %
& \R{0.9724 (+0.01)} %
\\
\hline             
\end{tabular}
\end{center}
\end{table*}

\begin{figure*}[!htbp]
	\captionsetup{font=small}
	
	\centering
	\scriptsize
	
	\renewcommand{\h}{0.105}
	\renewcommand{\wa}{0.12}
	\newcommand{\wb}{0.16}
	\renewcommand{\g}{-0.7mm}
	\renewcommand{\tabcolsep}{1.8pt}
	\renewcommand{\arraystretch}{1}
	\resizebox{1.00\linewidth}{!} {
		\begin{tabular}{cc}
			
			\renewcommand{\name}{figures/deblur/00034_}
			\renewcommand{\h}{0.12}
			\renewcommand{\w}{0.2}
			\begin{tabular}{cc}
				\begin{adjustbox}{valign=t}
					\begin{tabular}{c}%
					\includegraphics[trim={336 0 0 0 },clip, width=0.354\textwidth]{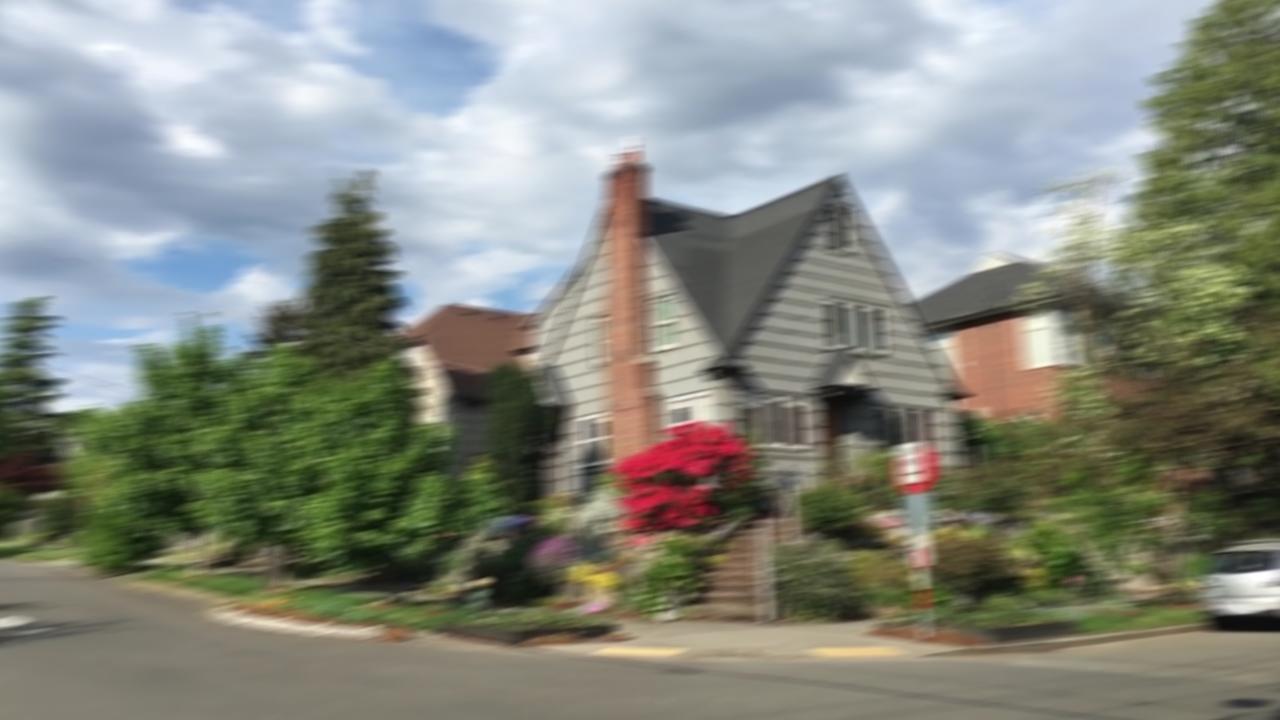}\\
						Frame 00034, Clip IMG\_0030, DVD~\cite{su2017dvddeblur} 
					\end{tabular}
				\end{adjustbox}
				\begin{adjustbox}{valign=t}
					\begin{tabular}{cccccc}
						\includegraphics[trim={830 100 285 530},clip,height=\h \textwidth, width=\w \textwidth]{\name LQ.jpg} \hspace{\g}  &
						\includegraphics[trim={830 325 285 530},clip,height=\h \textwidth, width=\w \textwidth]{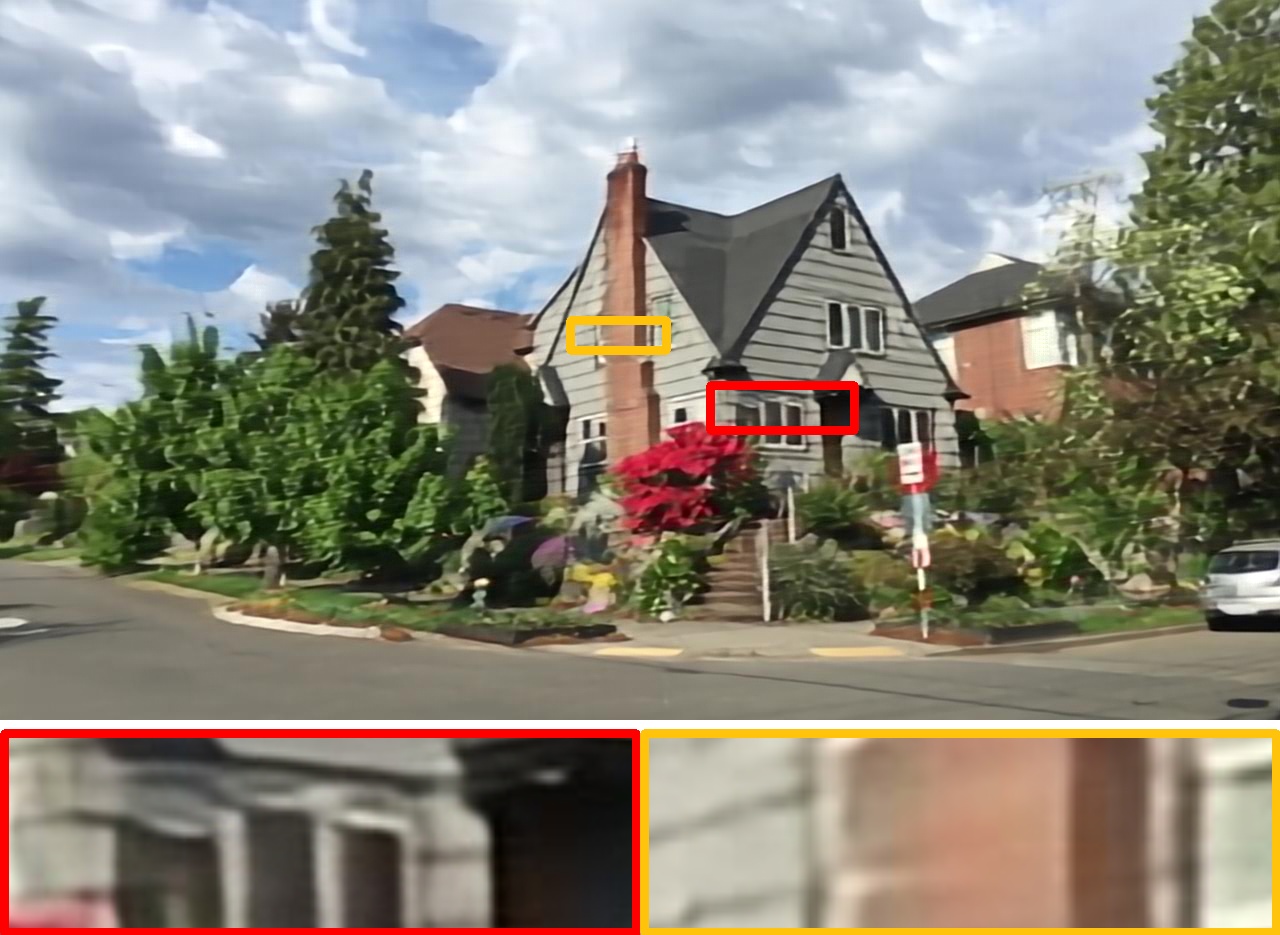} \hspace{\g} &
						\includegraphics[trim={830 325 285 530},clip,height=\h \textwidth, width=\w \textwidth]{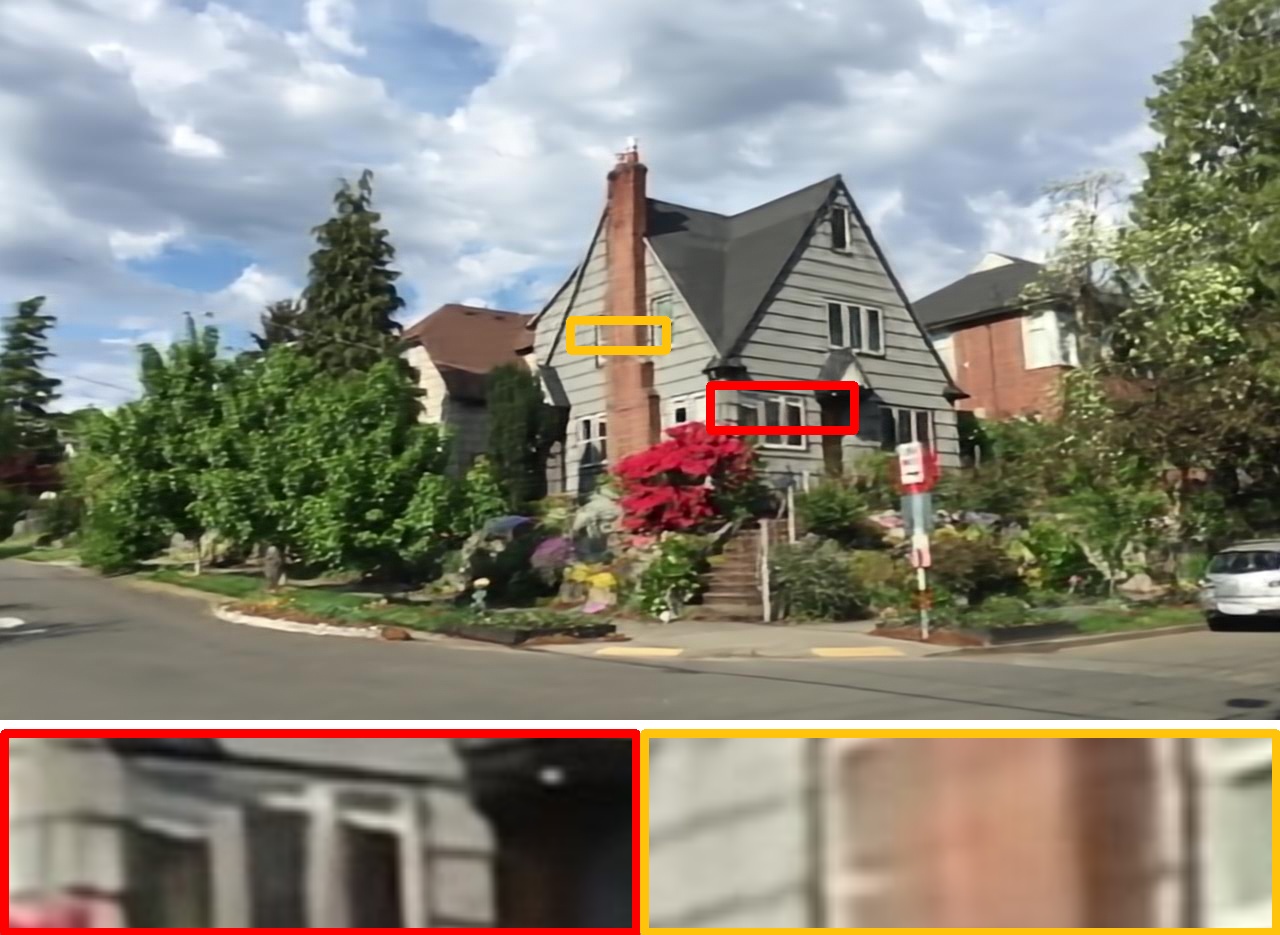} \hspace{\g} & \includegraphics[trim={830 325 285 530},clip,height=\h \textwidth, width=\w \textwidth]{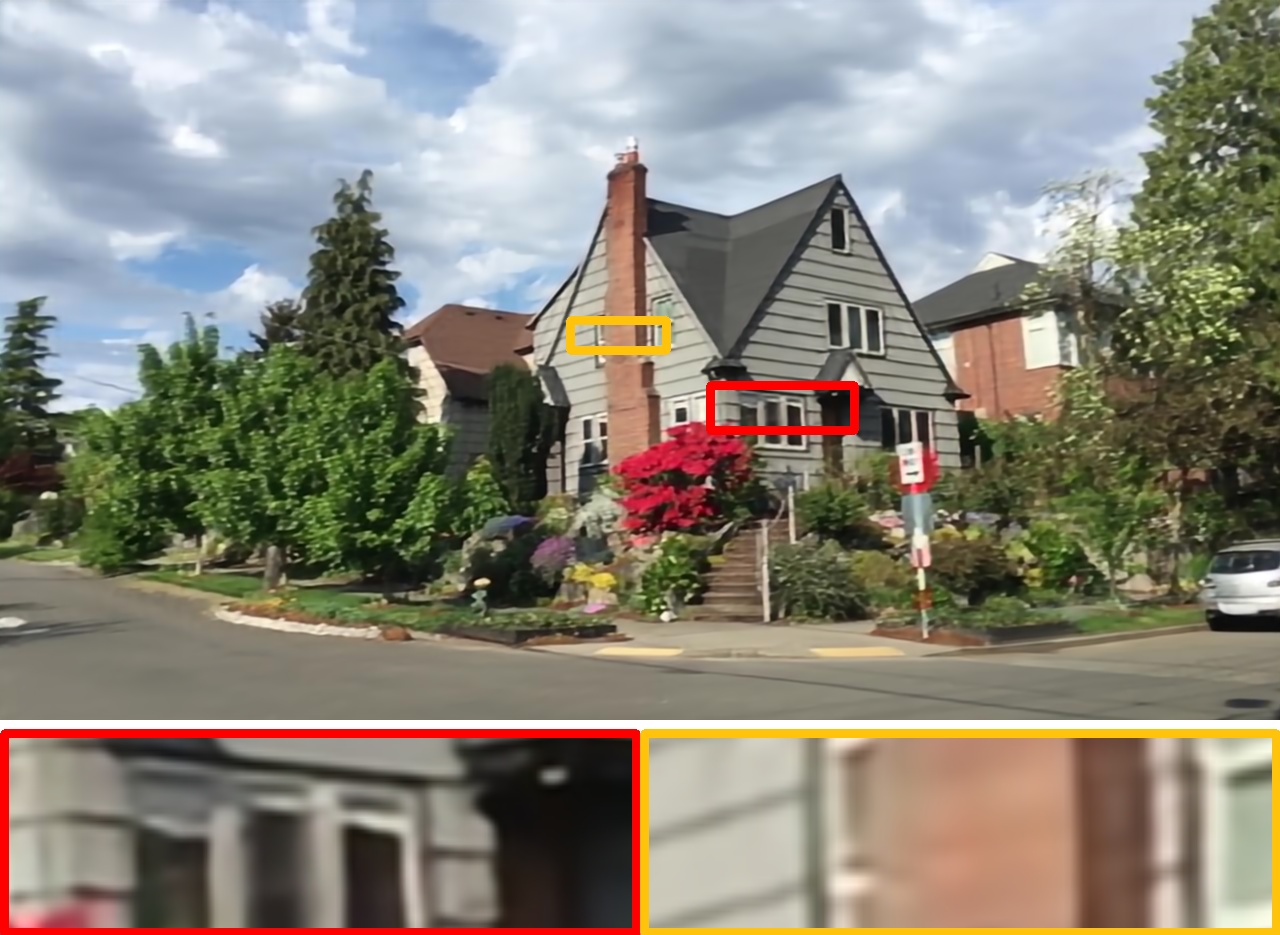} \hspace{\g} 
						\\
						LQ   & DBN~\cite{su2017dvddeblur}  & STFAN~\cite{zhou2019spatio} & TSP~\cite{pan2020tspdeblur} 
						\\
						\vspace{-1.5mm}
						\\
						\includegraphics[trim={830 100 285 530},clip,height=\h \textwidth, width=\w \textwidth]{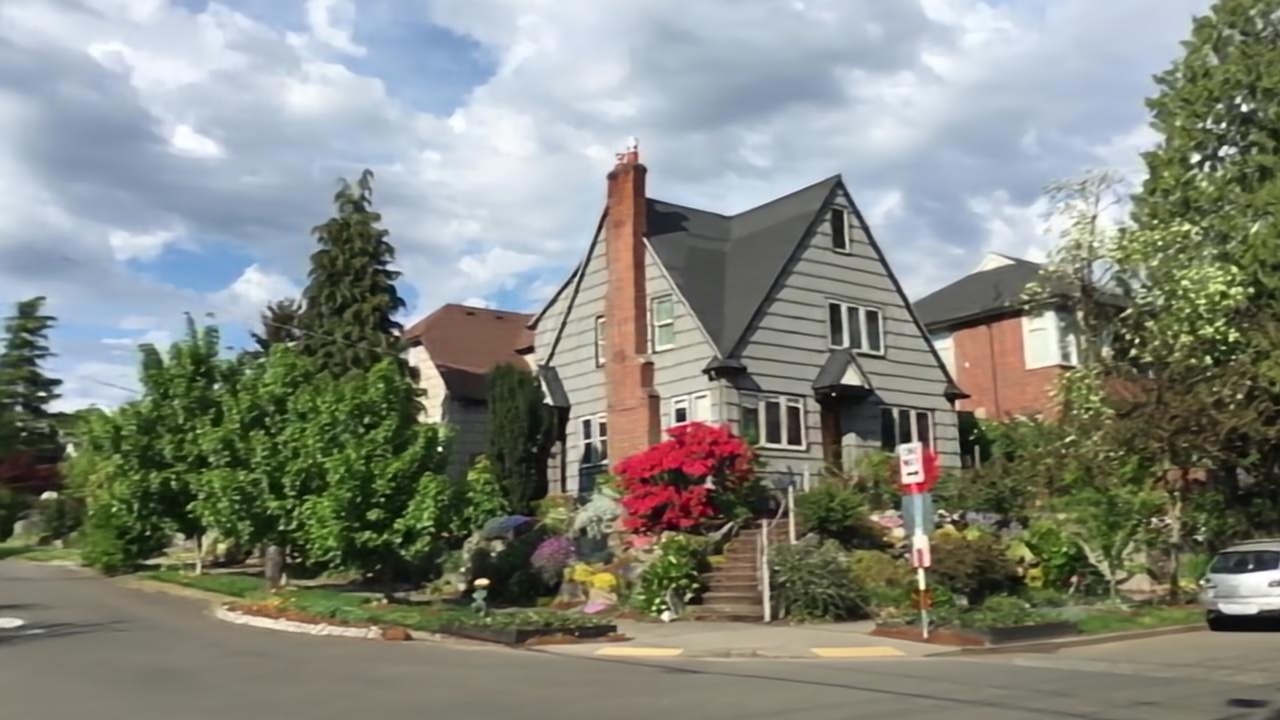} &
						\includegraphics[trim={830 325 285 530},clip,height=\h \textwidth, width=\w \textwidth]{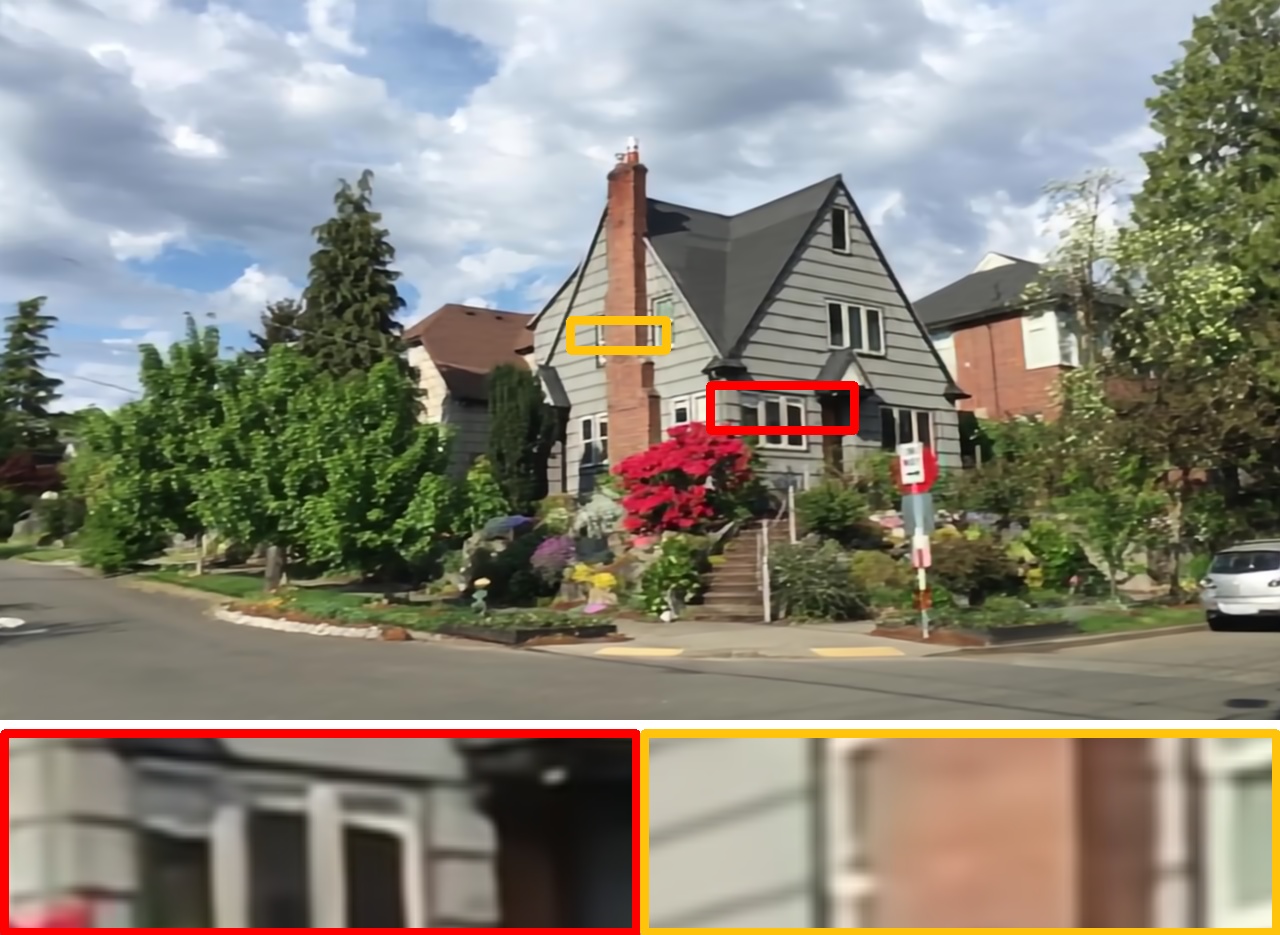} \hspace{\g} &
						\includegraphics[trim={830 100 285 530},clip,height=\h \textwidth, width=\w \textwidth]{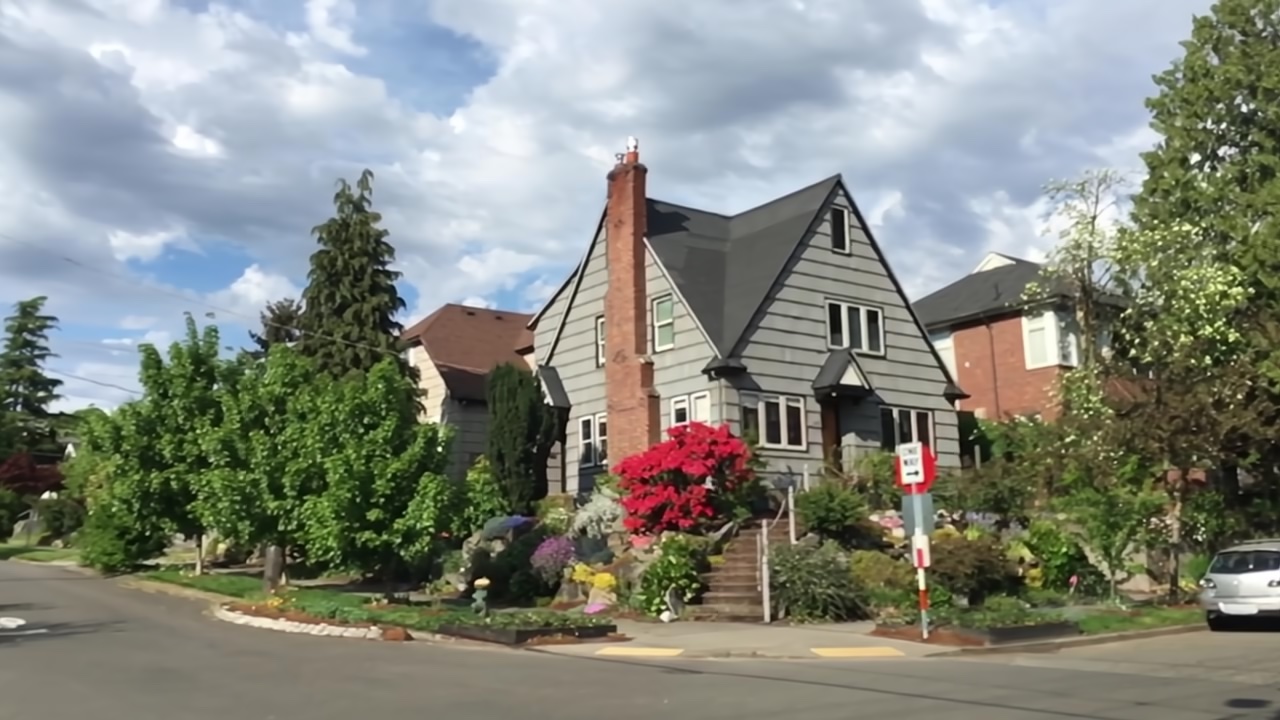}
						\hspace{\g} &		
						\includegraphics[trim={830 100 285 530},clip,height=\h \textwidth, width=\w \textwidth]{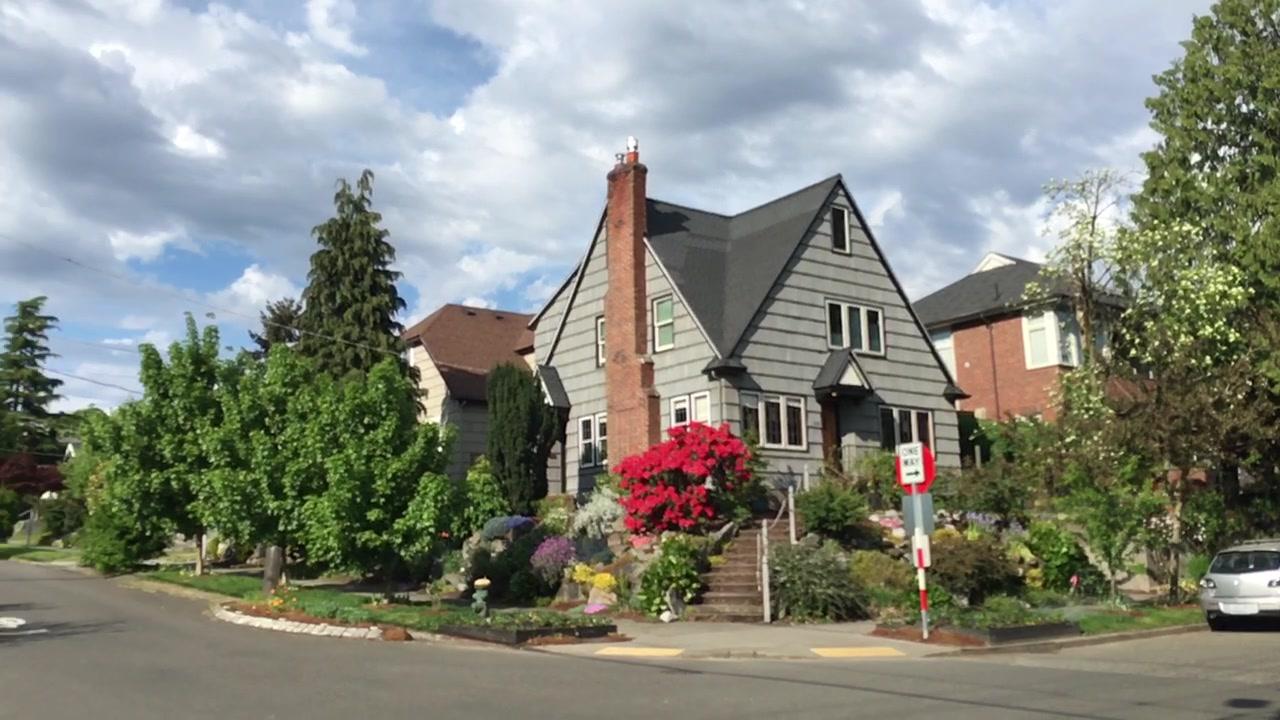} 
						\\ 
					     PVDNet~\cite{son2021recurrent} &
						ARVo~\cite{li2021arvo} &
						\textbf{VRT} (ours) &
						GT
						\\
					\end{tabular}
				\end{adjustbox}
			\end{tabular}
			
		\end{tabular}
	}
	\\ \vspace{1mm}
	\resizebox{1.00\linewidth}{!} {
		\begin{tabular}{cc}
			
			\renewcommand{\name}{figures/deblur/000210_}
			\renewcommand{\h}{0.12}
			\renewcommand{\w}{0.2}
			\begin{tabular}{cc}
				\begin{adjustbox}{valign=t}
					\begin{tabular}{c}
		         	\includegraphics[trim={90 0 90 100},clip,height=0.27\textwidth, width=0.354\textwidth]{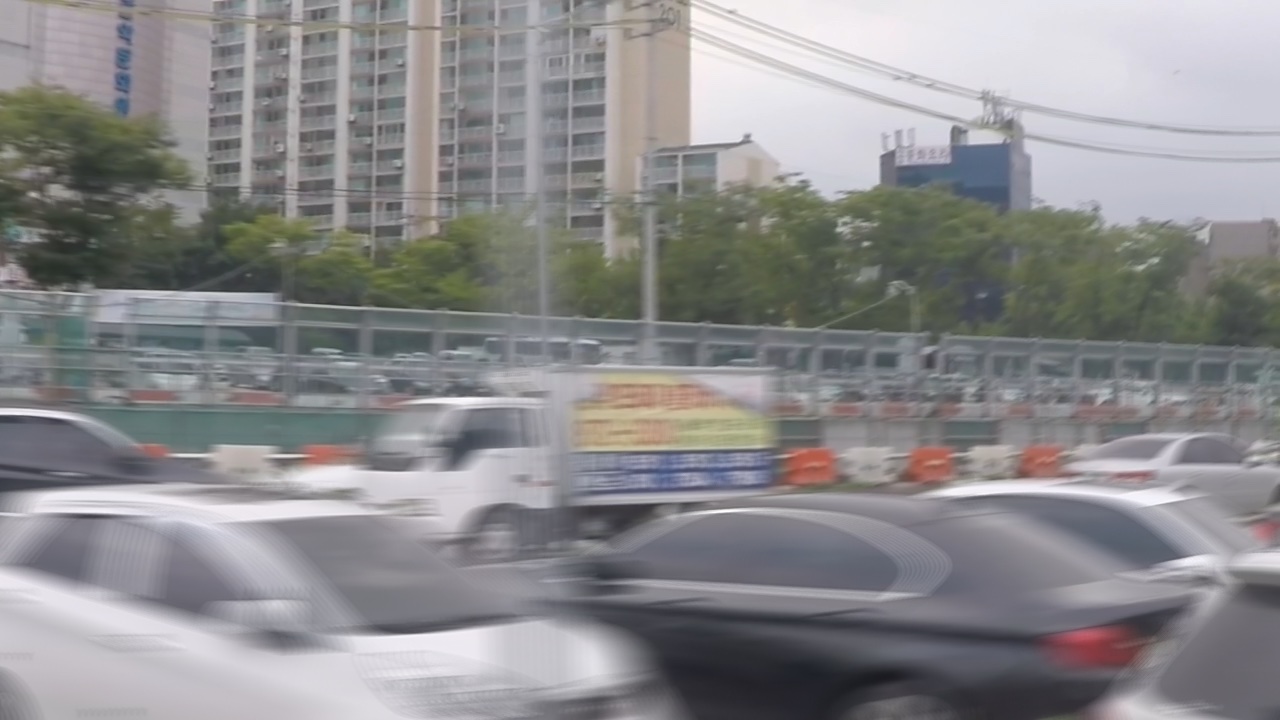}
						\\
						Frame 000210, Clip GOPRO410\_11\_00, GoPro~\cite{nah2017Gopro} 
					\end{tabular}
				\end{adjustbox}
				\begin{adjustbox}{valign=t}
					\begin{tabular}{cccccc}
						\includegraphics[trim={590 230 580 410 },clip,height=\h \textwidth, width=\w \textwidth]{\name LQ.jpeg} \hspace{\g} &
						\includegraphics[trim={590 230 580 410},clip,height=\h \textwidth, width=\w \textwidth]{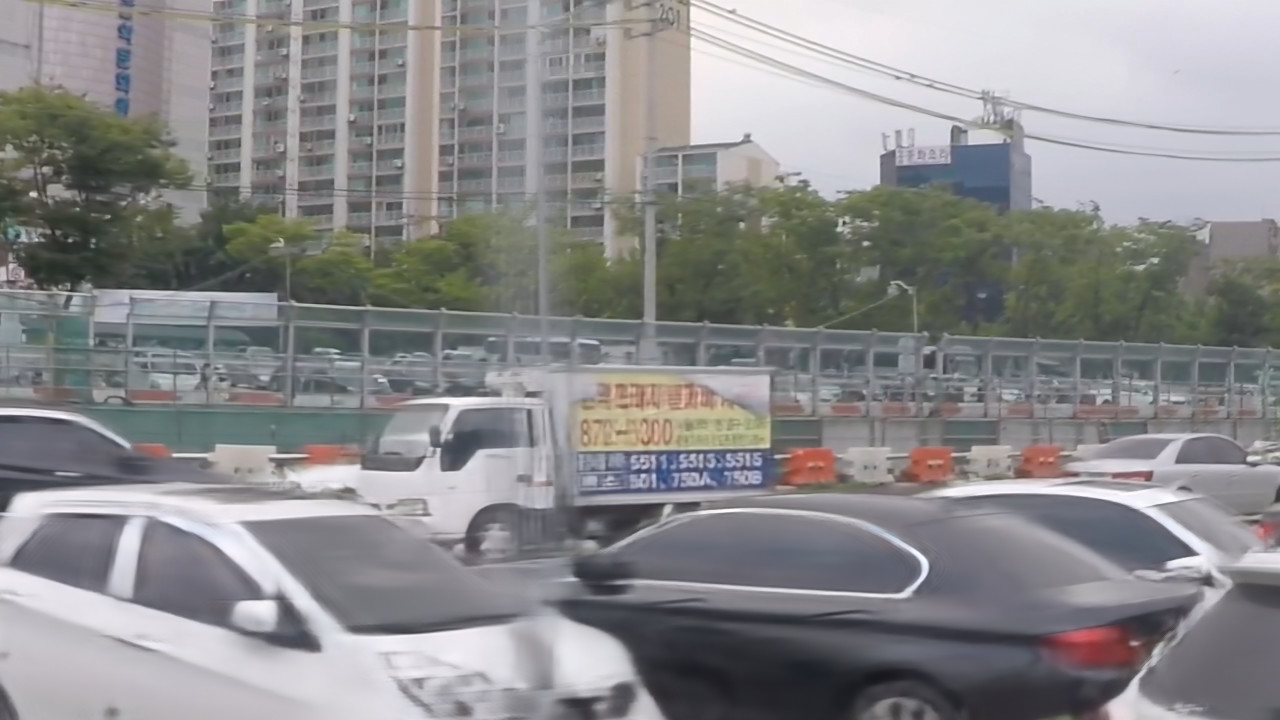} \hspace{\g} &
						\includegraphics[trim={590 230 580 410},clip,height=\h \textwidth, width=\w \textwidth]{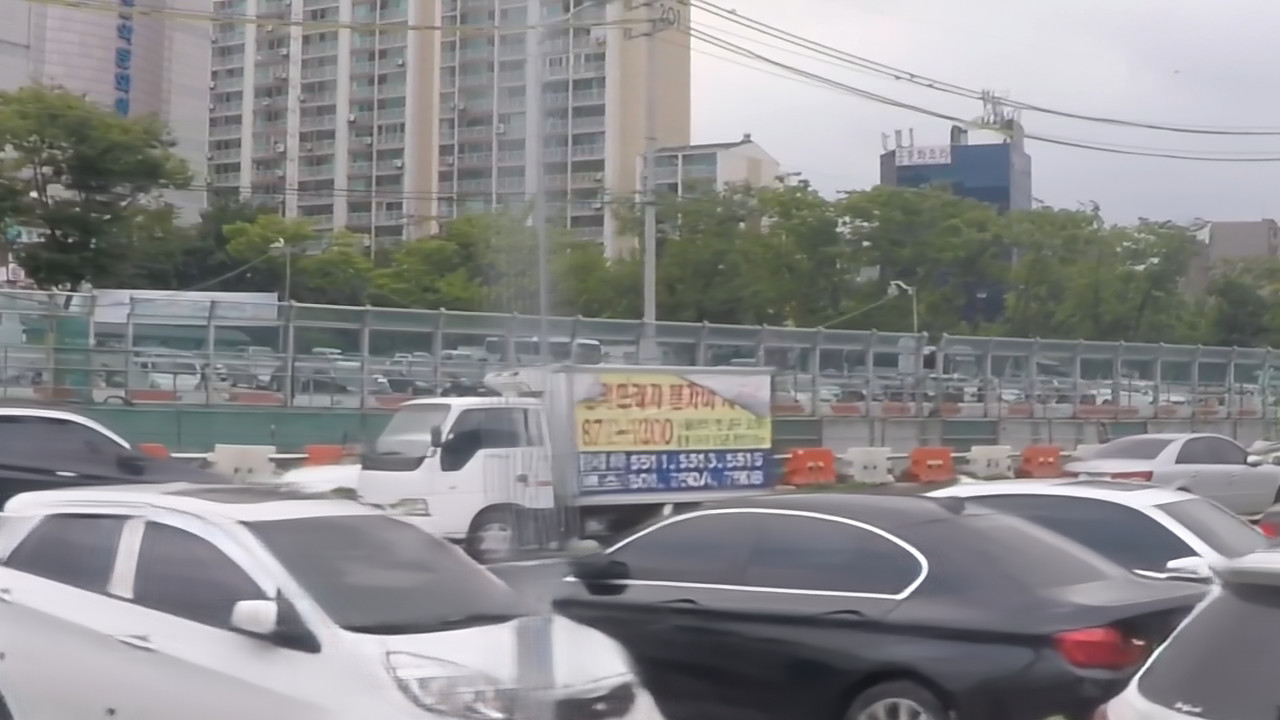} \hspace{\g} &
						\includegraphics[trim={590 230 580 410 },clip,height=\h \textwidth, width=\w \textwidth]{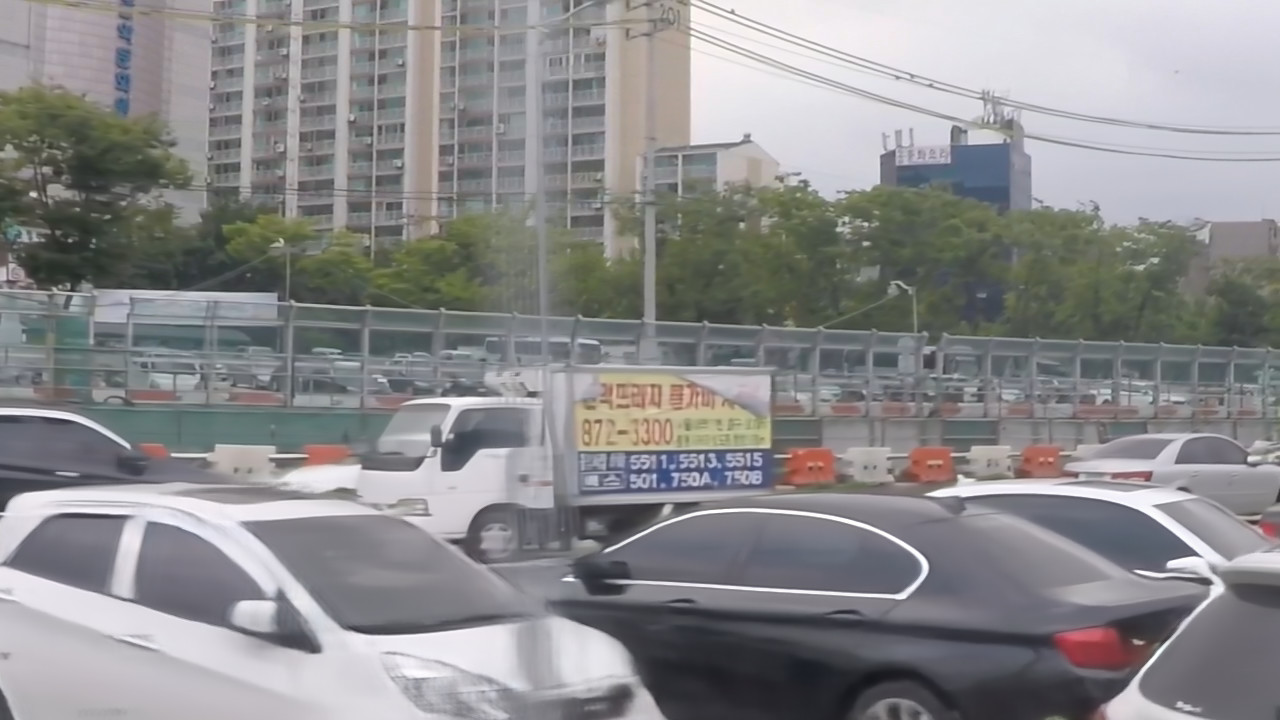} \hspace{\g} \\
						LQ  &
						SRN~\cite{tao2018scale}  &
						SAPHN~\cite{suin2020spatially} &
						MPRNet~\cite{zamir2021MPRNet}
						\\
						\vspace{-1.5mm}
						\\
						\includegraphics[trim={590 230 580 410 },clip,height=\h \textwidth, width=\w \textwidth]{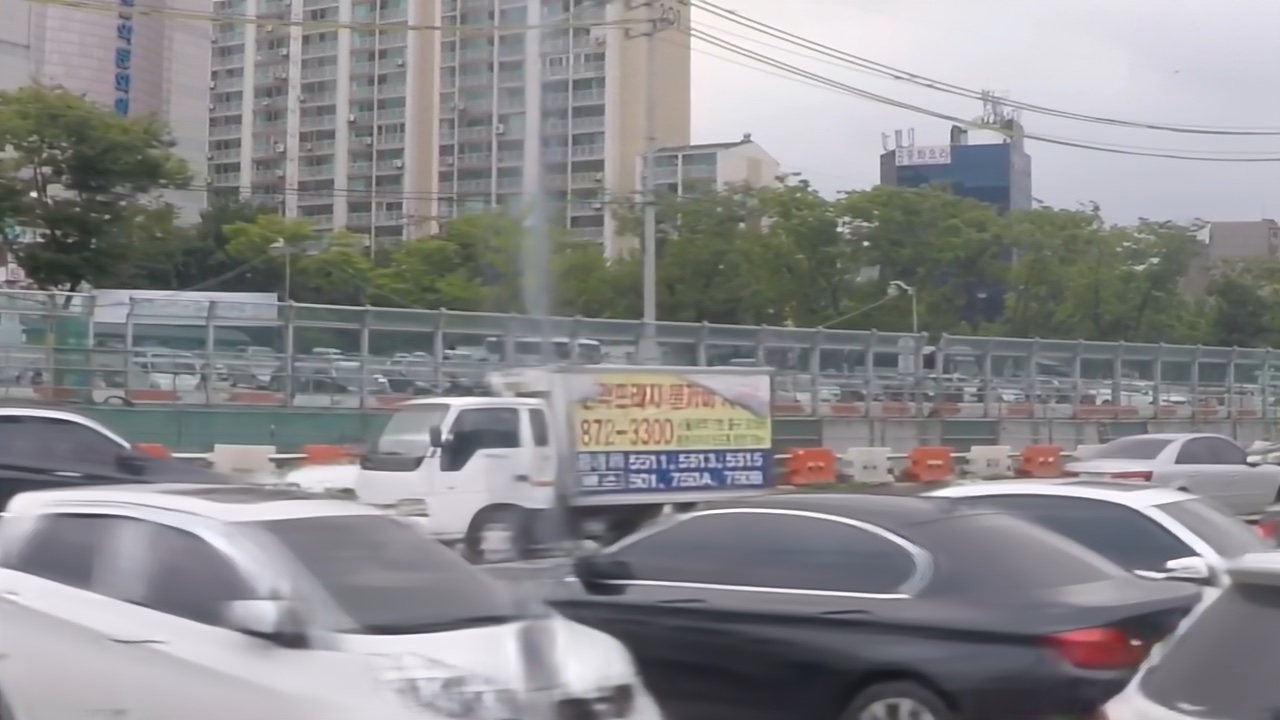} & 
						\includegraphics[trim={590 230 580 410 },clip,height=\h \textwidth, width=\w \textwidth]{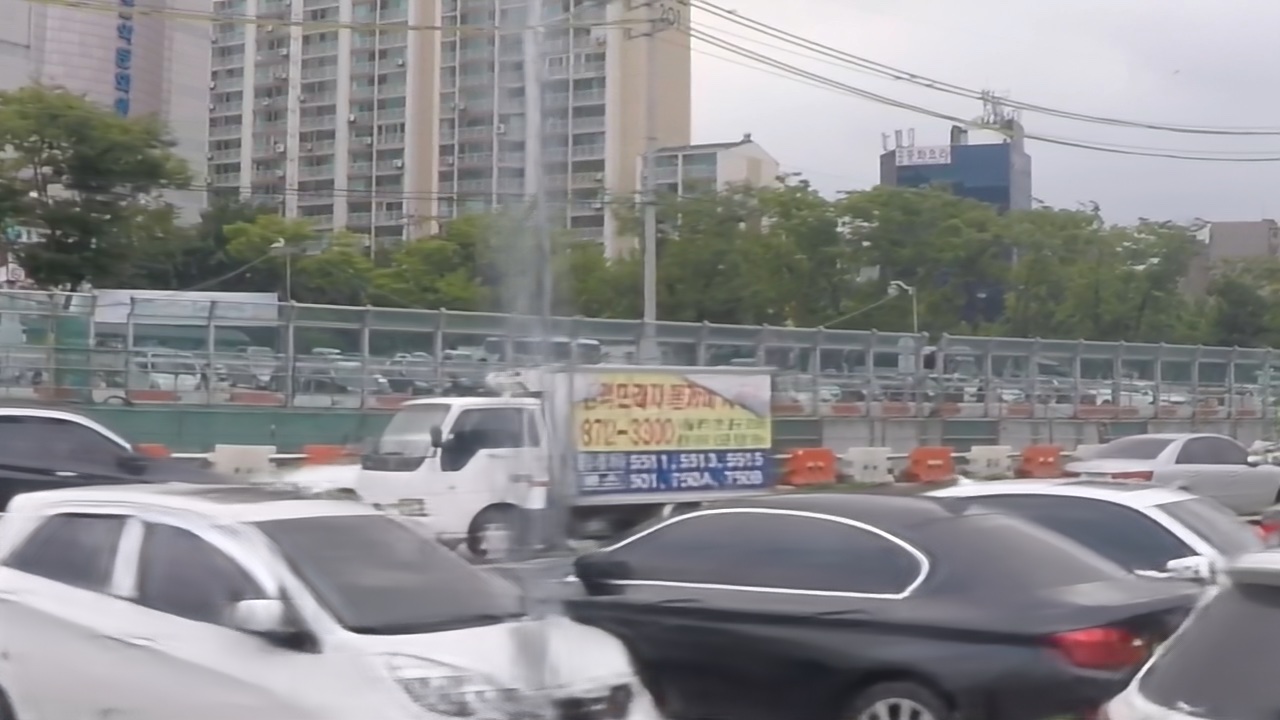} \hspace{\g} &
						\includegraphics[trim={590 230 580 410 },clip,height=\h \textwidth, width=\w \textwidth]{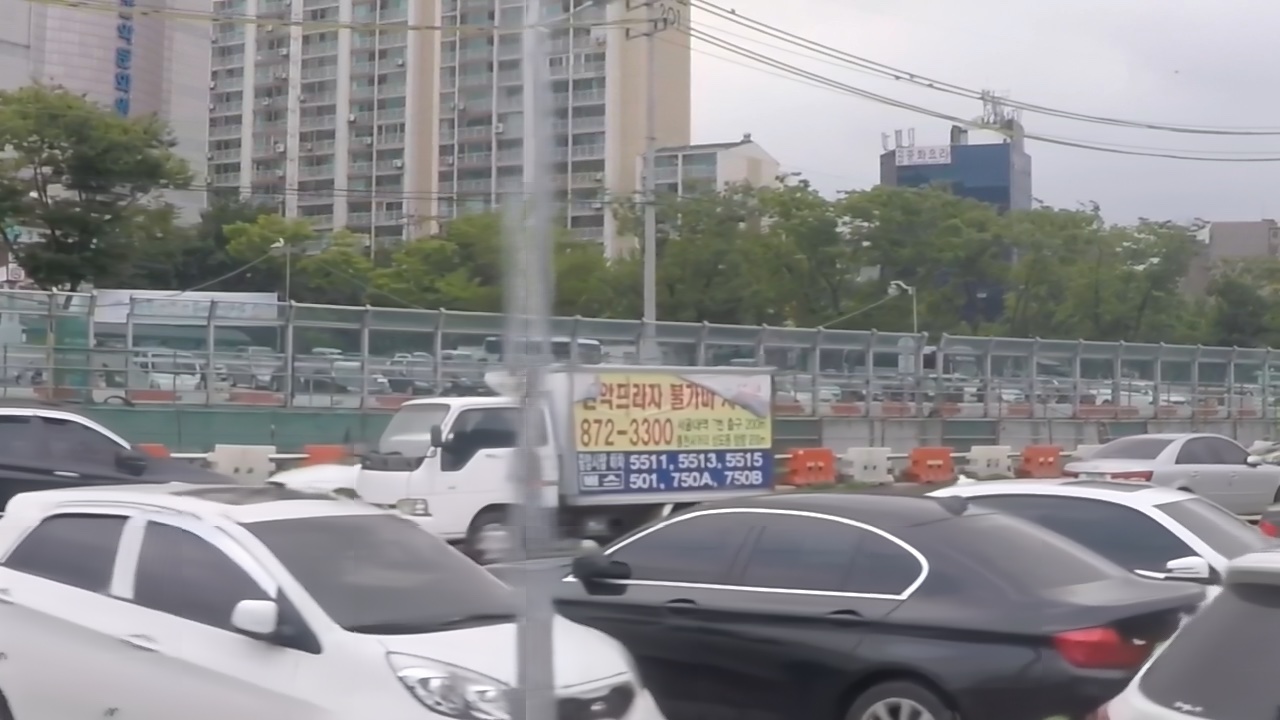}
						\hspace{\g} &		
						\includegraphics[trim={590 230 580 410 },clip,height=\h \textwidth, width=\w \textwidth]{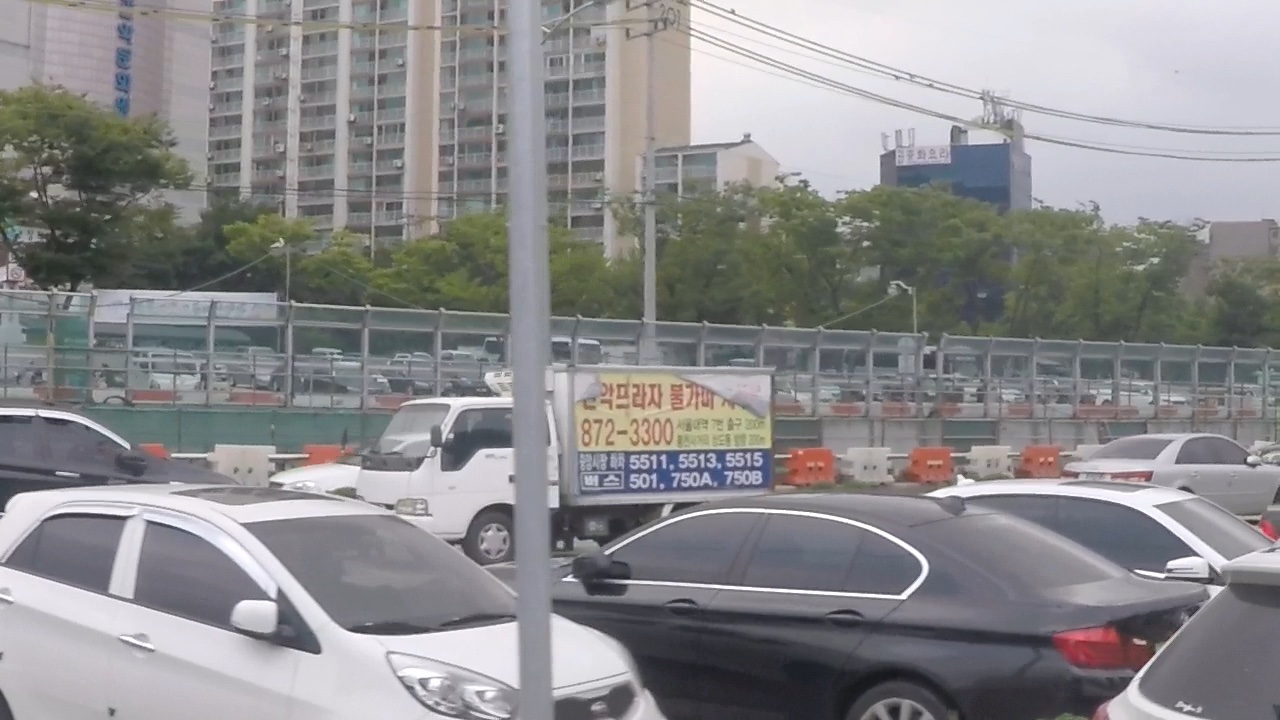} 
						\\ 
						TSP~~\cite{pan2020tspdeblur}  &  PVDNet~\cite{son2021recurrent} &
						\textbf{VRT} (ours) &
						GT
						\\
					\end{tabular}
				\end{adjustbox}
			\end{tabular}
			
		\end{tabular}
	}
	\vspace{-2mm}
	\caption{Visual comparison of \textbf{{video deblurring}} methods. Part of compared images are derived from~\cite{li2021arvo, zamir2021MPRNet}.}
	  \vspace{-1.2mm}
	\label{fig:deblur_quali}
\end{figure*}

\begin{table}[!t]\scriptsize
\center
\begin{center}
\caption{Quantitative comparison (average RGB channel PSNR/SSIM) with state-of-the-art methods for \textbf{{video deblurring}} on \textbf{REDS}~\cite{nah2019ntireREDS}. Best and second best results are in \R{red} and \B{blue} colors, respectively.}
\vspace{-2mm}
\label{tab:deblur_reds}
\begin{tabular}{|c|c|c||c|c|c|}
\hline
Method & 
\makecell{\scalebox{0.9}{DeepDeblur}\\\cite{nah2017Gopro}} &
\makecell{\scalebox{0.9}{SRN}\\\cite{tao2018scale}} &
\makecell{\scalebox{0.9}{DBN}\\\cite{su2017dvddeblur}} &
\makecell{\scalebox{0.9}{EDVR}\\\cite{wang2019edvr}} &
\makecell{\textbf{\scalebox{0.9}{VRT}}\scalebox{0.9}{ (ours)}}
\\
\hline
\hline
PSNR 
& 26.16
& 26.98
& 26.55
& \B{34.80}
& \R{36.79 (+1.99)}
\\
SSIM 
& 0.8249
& 0.8141
& 0.8066
& \B{0.9487}
& \R{0.9648 (+0.02)} %
\\
\hline             
\end{tabular}
\end{center}
\vspace{-1mm}
\end{table}

\subsection{Video Deblurring}
We conducts experiments on three different datasets for fair comparison with existing methods. Table~\ref{tab:deblur_dvd} shows the results on the DVD~\cite{su2017dvddeblur} dataset. It is clear that VRT achieves the best performance, outperforming the second best method ARVo by a remarkable improvement of 1.47dB and 0.0299 in terms of PSNR and SSIM. Related to the attention mechanism, GSTA designs a gated spatio-temporal attention mechanism, while ARVo calculates the correlation between pixel pairs for correspondence learning. However, both of them are based on CNN, achieving significantly worse performance compared with the Transformer-based VRT.
We also compare VRT on the GoPro~\cite{nah2017Gopro} and REDS~\cite{nah2019ntireREDS} datasets. VRT shows its superiority over other methods with significant PSNR gains of 2.15dB and 1.99dB. The total number of parameters of VRT is 18.3M, which is slightly smaller than EDVR (23.6M) and PVDNet (23.5M). The runtime is 2.2s per frame on $1280\times 720$ blurred videos. Notably, during evaluation, we do not use any pre-processing techniques such as sequence truncation and image alignment~\cite{pan2020tspdeblur, son2021recurrent}.

Fig.~\ref{fig:deblur_quali} shows the visual comparison of different methods. VRT is effective in removing motion blurs and restoring faithful details, such as the pole in the first example and characters in the second one. In comparison, other approaches fail to remove blurs completely and do not produce sharp edges.

\subsection{Video Denoising}
We also conduct experiments on video denoising to show the effectiveness of VRT. Following~\cite{tassano2019dvdnet, tassano2020fastdvdnet}, we train one non-blind model for noise level $\sigma\in [0,50]$ on the DAVIS~\cite{khoreva2018davis} dataset and test it on different noise levels. Table~\ref{tab:denoising_davis_set8} shows the superiority of VRT on two benchmark datasets over existing methods. Even though PaCNet~\cite{vaksman2021patch} trains different models separately for different noise levels, VRT still improves the PSNR by 0.82$\sim$2.16dB.

\subsection{Video Frame Interpolation}
To show the generalizability of our framework, we conduct experiments on video frame interpolation. Following~\cite{xu2019quadratic, kalluri2020flavr}, we train the model on Vimeo-90K~\cite{xue2019TOFlow-Vimeo-90K} for single frame interpolation and test it on quintuples generated from Vimeo-90K-T~\cite{xue2019TOFlow-Vimeo-90K}, UCF101~\cite{soomro2012ucf101} and DAVIS~\cite{khoreva2018davis}. As shown in Table~\ref{tab:vfi_vimeo_ucf_davis}, VRT achieves best or competitive performance on all datasets compared with it competitors, including those using depth maps or optical flows. As for the model size, VRT only has 9.9M parameters, which is much smaller than the recent best model FLAVR (42.4M).

\begin{table}[!t]\scriptsize
\center
\begin{center}
\caption{Quantitative comparison (average RGB channel PSNR) with state-of-the-art methods for \textbf{{video denoising}} on \textbf{DAVIS}~\cite{khoreva2018davis} and \textbf{Set8}~\cite{tassano2019dvdnet}. $\sigma$ is the additive white Gaussian noise level. Best and second best results are in \R{red} and \B{blue} colors, respectively.}
\vspace{-2mm}
\label{tab:denoising_davis_set8}
\begin{tabular}{|c|c|c|c|c|c|c|}
\hline
Dataset & $\sigma$ &
\makecell{\scalebox{0.9}{VLNB}\\\cite{arias2018video}} &
\makecell{\scalebox{0.9}{DVDnet}\\\cite{tassano2019dvdnet}} &
\makecell{\scalebox{0.9}{FastDVDnet}\\\cite{tassano2020fastdvdnet}} &
\makecell{\scalebox{0.9}{PaCNet}\\\cite{vaksman2021patch}} & %
\makecell{\textbf{\scalebox{0.9}{VRT}}\scalebox{0.9}{ (ours)}}
\\
\hline
\hline
\multirow{5}{*}{DAVIS}  
& 10
& 38.85
& 38.13
& 38.71
& 39.97 %
& \R{40.82 (+0.85)}
\\
& 20
& 35.68
& 35.70
& 35.77
& 36.82
& \R{38.15 (+1.33)}
\\
& 30
& 33.73
& 34.08
& 34.04
& 34.79
& \R{36.52 (+1.73)}
\\
& 40
& 32.32
& 32.86
& 32.82
& 33.34
& \R{35.32 (+1.98)}
\\
& 50
& 31.13
& 31.85
& 31.86
& 32.20
& \R{34.36 (+2.16)}
\\\hline\hline
\multirow{5}{*}{Set8}  
& 10
& 37.26
& 36.08
& 36.44
& 37.06
& \R{37.88 (+0.82)}
\\
& 20
& 33.72
& 33.49
& 33.43
& 33.94
& \R{35.02 (+1.08)}
\\
& 30
& 31.74
& 31.79
& 31.68
& 32.05
& \R{33.35 (+1.30)}
\\
& 40
& 30.39
& 30.55
& 30.46
& 30.70
& \R{32.15 (+1.45)}
\\
& 50
& 29.24
& 29.56
& 29.53
& 29.66
& \R{31.22 (+1.56)}\\
\hline             
\end{tabular}
\end{center}
\end{table}

\begin{table}[!t]\scriptsize
  \centering
    \caption{Quantitative comparison (average RGB channel PSNR) with state-of-the-art methods for \textbf{video frame interpolation} (single frame interpolation, $\times 2$) on \textbf{Vimeo-90K-T}~\cite{xue2019TOFlow-Vimeo-90K}, \textbf{UCF101}~\cite{soomro2012ucf101} and \textbf{DAVIS}~\cite{khoreva2018davis}. R, D and F means that the model uses RGB images, depth maps and optical flows, respectively.}
    \vspace{-2mm}
     \label{tab:vfi_vimeo_ucf_davis}
  \begin{tabular}{|c|c|c|c|c|c|} 
    \hline 
    Method & Inputs & \makecell{\scalebox{0.9}{Vimeo-90K-T}\\\cite{xue2019TOFlow-Vimeo-90K}}  & \makecell{\scalebox{0.9}{UCF101}\\\cite{soomro2012ucf101}} & \makecell{\scalebox{0.9}{DAVIS}\\\cite{khoreva2018davis}}\\
    \hline
    \hline
    DAIN~\cite{bao2019depth}            &   R+D+F  &   33.35/0.945    &       31.64/0.957     &   26.12/0.870    \\
    QVI~\cite{xu2019quadratic}          &   R+F       &   35.15/{0.971}    &      {32.89}/{0.970}     &   {27.17}/{0.874}     \\
    DVF~\cite{liu2017video}             &   R             &   27.27/0.893     &      28.72/0.937     &    22.13/0.800      \\
    SepConv~\cite{niklaus2017video}     &   R             &   33.60/0.944     &      31.97/0.943     &    26.21/0.857      \\
    CAIN~\cite{choi2020channel}         &   R             &   33.93/0.964     &      32.28/0.965     &    26.46/0.856      \\
    SuperSloMo~\cite{jiang2018super}    &   R             &   32.90/0.957      &      32.33/0.960     &    25.65/0.857      \\
    {BMBC}~\cite{park2020bmbc} & R & 34.76/0.965 & {32.61}/0.955 & 26.42/{0.868} \\
    {AdaCoF}~\cite{lee2020adacof} & R & {35.40}/{0.971} & {32.71}/0.969 & 26.49/0.866 \\
    FLAVR~\cite{kalluri2020flavr}        &   R             &  \B{ 36.25/0.975}      &      \R{33.31/0.971}     &    \B{27.43/0.874}   \\
    \textbf{VRT} (ours)      &   R      & \R{36.53/0.977} & \B{33.30/0.970} &  \R{27.88/0.889} \\
    \hline
  \end{tabular}
\end{table}

\subsection{Space-Time Video Super-Resolution}
With the pretrained models on video SR (VSR) and video frame interpolation (VFI), we directly test VRT on space-time video super-resolution by cascading VRT models in two ways: VFI followed by VSR, or VSR followed by VFI. As shown in Table~\ref{tab:stvsr_vimeo_vid4}, compared with existing methods, VRT provides a strong baseline for space-time video super-resolution, even though it serves as a two-stage model and is not specifically trained for this task. In particular, it improves the PSNR by 1.03dB on the Vid4 dataset.

\begin{table}[!t]\scriptsize
\caption{Quantitative comparison (average Y channel PSNR) with state-of-the-art methods for \textbf{space-time video super-resolution} (time: $\times 2$, space: $\times 4$) on \textbf{Vid4}~\cite{liu2013bayesianVid4} and \textbf{Vimeo-90K-T}~\cite{xue2019TOFlow-Vimeo-90K}. \cite{jiang2018super}, \cite{niklaus2017video} and \cite{bao2019depth} are frame interpolation methods SuperSloMo, SepConv and DAIN, respectively. Note that the proposed VRT is not trained on this task.}
 \vspace{-2mm}
\label{tab:stvsr_vimeo_vid4}
\scalebox{0.9}{
\begin{tabular}{|c|c|c|c|c|}
\hline
\makecell{VFI+VSR\\Methods} & \makecell{\scalebox{1}{Vid4}\\\cite{liu2013bayesianVid4}} & \makecell{\scalebox{0.8}{Vimeo-Fast}\\\cite{xue2019TOFlow-Vimeo-90K}} & \makecell{\scalebox{0.8}{Vimeo-Medium}\\\cite{xue2019TOFlow-Vimeo-90K}} & \makecell{\scalebox{0.8}{Vimeo-Slow}\\\cite{xue2019TOFlow-Vimeo-90K}}\\
\hline \hline
\cite{jiang2018super}+Bicubic                                                                   &   22.84/0.5772 &   	31.88/0.8793 &   	29.94/0.8477 &   	28.37/0.8102 \\
\cite{jiang2018super}+RCAN \cite{zhang2018rcan}                                                                  & 23.80/0.6397      & 34.52/0.9076         & 32.50/0.8884          & 30.69/0.8624   \\
\cite{jiang2018super}+RBPN \cite{haris2019recurrent}                                                                  & 23.76/0.6362      & 34.73/0.9108    &      32.79/0.8930	 & 30.48/0.8584  \\
\cite{jiang2018super}+EDVR \cite{wang2019edvr}                                                                  & 24.40/0.6706	  & 35.05/0.9136  & 	33.85/0.8967  & 	30.99/0.8673 \\ 
\cite{niklaus2017video}+Bicubic                                                                   &   23.51/0.6273 &   	32.27/0.8890 &   	30.61/0.8633 &   	29.04/0.8290                \\
\cite{niklaus2017video}+RCAN \cite{zhang2018rcan}                                                                  & 24.92/0.7236      & 34.97/0.9195         & 33.59/0.9125          & 32.13/0.8967   \\
\cite{niklaus2017video}+RBPN \cite{haris2019recurrent}                                                                  & 26.08/0.7751      & 35.07/0.9238         &       34.09/0.9229           & 32.77/0.9090       \\
\cite{niklaus2017video}+EDVR \cite{wang2019edvr}                                                                  & 25.93/0.7792      & 35.23/0.9252         & 34.22/0.9240            & 32.96/0.9112    \\
\cite{bao2019depth}+Bicubic                                                                &    23.55/0.6268  &   	32.41/0.8910  &   	30.67/0.8636  &   	29.06/0.8289    \\
\cite{bao2019depth}+RCAN \cite{zhang2018rcan}                                                                  & 25.03/0.7261      & 35.27/0.9242         & 33.82/0.9146          & 32.26/0.8974    \\
\cite{bao2019depth}+RBPN \cite{haris2019recurrent}                                                                  & 25.96/0.7784      & 35.55/0.9300         & 34.45/0.9262          & 32.92/0.9097      \\
\cite{bao2019depth}+EDVR \cite{wang2019edvr}                                                                  & {26.12}/{0.7836}        & {35.81}/{0.9323}       & {34.66}/{0.9281}            & {33.11}/{0.9119}               \\ %
ZSM~\cite{xiang2020zooming}  & {26.31}/{0.7976}       &  {36.81}/{0.9415}          &  {35.41}/{0.9361}           &  {33.36}/{0.9138}      \\
STARnet~\cite{haris2020space}  & 26.06/0.8046 & 36.19/0.9368 & 34.86/0.9356 & 33.10/0.9164  \\
TMNet~\cite{xu2021temporal} & 26.43/0.8016 & \R{37.04/0.9435} & 35.60/0.9380 & 33.51/0.9159\\
RSTT~\cite{geng2022rstt} & 26.43/0.7994 & 36.80/0.9403 & \B{35.66/0.9381} & 33.50/0.9147 \\
\textbf{VRT} (VFI+VSR) & \B{26.59/0.8014} &  36.56/0.9372  & 35.28/0.9343 & \B{33.75/0.9204} \\
\textbf{VRT} (VSR+VFI) & \R{27.46/0.8392} & \B{36.98/0.9439}   & \R{36.01/0.9434} &  \R{34.01/0.9236} \\

\hline
\end{tabular}}
\end{table}

\subsection{Ablation Study}
For ablation study, we set up a small version of VRT as the baseline model by halving the layer and channel numbers. All models are trained on Vimeo-90K~\cite{xue2019TOFlow-Vimeo-90K} for bicubic video SR ($\times 4$) and tested it on Vid4~\cite{liu2013bayesianVid4}. 

\vspace{-0.45cm}
\paragraph{Impact of multi-scale architecture \& parallel warping.}
Table~\ref{tab:ablation_mutil_scale} shows the ablation study on the multi-scale architecture and parallel warping. When the number of model scales is reduced, the performance drops gradually, even though the computation burden becomes heavier. This is expected because multi-scale processing can help the model utilize information from a larger area and deal with large motions between frames. Besides, parallel warping also helps, bringing an improvement of 0.17dB.

\vspace{-0.45cm}
\paragraph{Impact of temporal mutual self attention.}
To test the effectiveness of mutual and self attention in TMSA, we conduct ablation study in Table~\ref{tab:ablation_mama}. When we replace mutual attention with self attention (\ie, two self attentions) or only use one self attention, the performance drops by 0.11$\sim$0.17dB. One possible reason is that the model may be more focused on the reference frame rather than on the supporting frame during the computation of attention maps. In contrast, using the mutual attention can help the model to explicitly attend to the supporting frame and benefit from feature fusion. In addition, we can find that only using mutual attention is not enough. This is because mutual attention cannot preserve information of reference frames.

\vspace{-0.45cm}
\paragraph{Impact of attention window size.}
We conduct ablation study in Table~\ref{tab:ablation_temporal_size} to investigate the impact of attention window size in the last few TMSAs of each scale. When the temporal window size increases from 1 to 2, the performance only improves slightly, possibly due to the fact that previous TMSA layers can already make good use of neighboring two-frame information. When the size is increased to 8, we can see an obvious improvement of 0.18dB. As a result, we use the window size of $8\times 8\times 8$ for those layers.

\begin{table}[t]
\captionsetup{font=small}%
\scriptsize
\center
\begin{center}
\caption{Ablation study on multi-scale architecture and parallel warping. Given an input of spatial size $64\times 64$, the corresponding feature sizes of each scale are shown in brackets. When some scales are removed, we add more layers to the rest scales to keep similar model size.}
\vspace{-2mm}
\label{tab:ablation_mutil_scale}
\begin{tabular}{|c|c|c|c|c|c|c|}
\hline
\renewcommand{\arraystretch}{0.4}
 \makecell{1\\($64\times 64$)} & \makecell{2\\($32\times 32$)} & \makecell{3\\($16\times 16$)} & \makecell{4\\($8\times 8$)} & \makecell{Parallel\\warping} &  PSNR\\\hline
 \checkmark &   &   &   & \checkmark & 27.13 \\
 \checkmark  & \checkmark  &  &  & \checkmark & 27.20 \\
 \checkmark  & \checkmark  & \checkmark  &  &\checkmark &  27.25 \\
 \checkmark & \checkmark & \checkmark & \checkmark & & 27.11\\\hline
 \checkmark  & \checkmark  & \checkmark  & \checkmark  &\checkmark &  {27.28} \\\hline
\end{tabular}
\end{center}
\end{table}

\begin{table}[t]
\captionsetup{font=small}%
\scriptsize
\center
\begin{center}
\caption{Ablation study on temporal mutual self attention.}
\vspace{-2mm}
\label{tab:ablation_mama}
\begin{tabular}{|c|c|c|c|c|c|}
\hline
Attention 1 &  Self Attn. & - & Mutual Attn. & Mutual Attn. \\\hline
Attention 2 & Self Attn. & Self Attn. & - & Self Attn.\\\hline
PSNR & 27.17 & 27.11 & 26.92 & 27.28\\
\hline
\end{tabular}
\end{center}
\end{table}

\begin{table}[t]
\captionsetup{font=small}%
\scriptsize
\center
\begin{center}
\caption{Ablation study on attention window size (frame $\times$ height $\times$ width).}%
\vspace{-2mm}
\label{tab:ablation_temporal_size}
\begin{tabular}{|c|c|c|c|c|c|}
\hline
\renewcommand{\arraystretch}{0.4}
Window Size  & $1\times 8\times 8$ &  $2\times 8\times 8$ & $4\times 8\times 8$ & $8\times 8\times 8$\\\hline
PSNR  & 27.10 & 27.13 & 27.18 &  27.28\\
\hline
\end{tabular}
\end{center}
\vspace{-0.1cm}
\end{table}

\section{Conclusion}
In this paper, we proposed the Video Restoration Transformer (VRT) for video restoration. Based on a multi-scale framework, it jointly extracts, aligns, and fuses information from different frames at multiple resolutions by two kinds of modules: multiple temporal mutual self attention (TMSA) and parallel warping. More specifically, TMSA is composed of mutual and self attention. Mutual attention allows joint implicit flow estimation and feature warping, while self attention is responsible for feature extraction. Parallel warping is also used to further enhance feature alignment and fusion. Extensive experiments on various benchmark datasets show that VRT 
leads to significant performance gains (up to 2.16dB) for video restoration, including video super-resolution, video deblurring, video denoising, video frame interpolation and space-time video super-resolution.

\vspace{0.8cm}
\noindent\textbf{Acknowledgements}~~ This work was partially supported by the ETH Zurich Fund (OK), a Huawei Technologies Oy (Finland) project, the China Scholarship Council and an Amazon AWS grant. Thanks Dr. Gurkirt Singh for insightful discussions. Special thanks goes to Yijue Chen.

{\small
\bibliographystyle{ieee_fullname}
\bibliography{superresolution.bib}
}

\end{document}